\documentclass[runningheads]{llncs}
\usepackage{graphicx}
\usepackage{subfig}
\usepackage{xcolor}
\usepackage[export]{adjustbox} 
\usepackage{stackengine}
\usepackage{array}
\usepackage{arydshln}
\usepackage{multirow}
\usepackage{pgfplots}
\usepackage{pgfplotstable}
\usepackage{amsmath}
\usepackage{hyperref}
\usepackage{standalone} 

\usepackage{graphicx}
\usepackage{tikz}
\usetikzlibrary{shapes.geometric,shadows,arrows,arrows.meta,fit,calc,positioning,decorations.markings,backgrounds}
\pgfdeclarelayer{background}
\pgfdeclarelayer{foreground}
\pgfsetlayers{background,main,foreground} 
\usetikzlibrary{shadows}
\usepackage{pgfplots}
\usepackage{pgfplotstable}
\usepackage{amsmath}
\usetikzlibrary{shapes.geometric}
\definecolor{cbblue}{RGB}{179,205,227}
\definecolor{cbred}{RGB}{251,180,174}
\definecolor{cbgreen}{RGB}{204,235,197}

\definecolor{mycolor}{RGB}{153, 0, 204}
\definecolor{mergefill}{RGB}{230, 204, 255}
\definecolor{sourcecol}{RGB}{0, 64, 255}
\definecolor{targetcol}{RGB}{255, 51, 51}

\usepackage{ifthen}
\newboolean{long}   
\setboolean{long}{false}

\newcommand{\titleZ}{4.5}

\tikzstyle{fig} = [rectangle, minimum width=0.5cm, minimum height=0.5cm]
\tikzstyle{EL} = [ellipse,minimum width=5cm,minimum height=3cm,fill=cbblue]
\tikzstyle{arrow} = [thick,->,>=stealth,bend left]

\tikzstyle{title_style} = [rectangle, minimum width=0.5cm, minimum height=0.5cm,draw=white,font=\LARGE]

\newcommand{\Lreg}{\mathcal{L}_{sim}}
\newcommand{\Lseg}{\mathcal{L}_{seg}}
\newcommand{\Lsmooth}{\mathcal{L}_{smooth}}	
\newcommand{\Ljac}{\mathcal{L}_{jac}}

\newcommand{\height}{3cm}

\begin{document}
%

\title{Exploring Deep Registration Latent Spaces}
\titlerunning{Exploring Deep Registration Latent Spaces}
%
\author{Th\'eo Estienne\inst{1,2} \and Maria Vakalopoulou\inst{1} \and Stergios Christodoulidis\inst{1} \and Enzo Battistella\inst{1,2} \and Théophraste Henry \inst{2} \and Marvin Lerousseau\inst{1,2}  \and Amaury Leroy\inst{2,4} \and
Guillaume Chassagnon \inst{3} \and Marie-Pierre Revel \inst{3} \and Nikos Paragios\inst{4} \and Eric Deutsch\inst{2}}
\index{Estienne, Théo} 
\index{Vakalopoulou, Maria}
\index{Christodoulidis, Stergios}
\index{Battistella, Enzo}
\index{Henry, Théophraste}
\index{Lerousseau, Marvin}
\index{Leroy, Amaury}
\index{Chassagnon, Guillaume}
\index{Revel, Marie-Pierre}
\index{Paragios, Nikos}
\index{Deutsch, Eric}

%
\institute{Université Paris-Saclay, CentraleSupélec, Mathématiques et Informatique pour la Complexité et les Systèmes, Inria Saclay, 91190, Gif-sur-Yvette, France. \email{theo.estienne@centralesupelec.fr} \and  Université Paris-Saclay, Institut Gustave Roussy, Inserm, Radiothérapie Moléculaire et Innovation Thérapeutique, 94800, Villejuif, France.\and Departement de Radiology, Hôpital Cochin, AP-HP Centre, Université de Paris, 27 Rue du Faubourg Saint-Jacques, 75014 Paris, France\and TheraPanacea, P\'epiniere Sant\'e Cochin, Paris, France.}

\authorrunning{Estienne T. et al.}

%


\maketitle              
\begin{abstract}
Explainability of deep neural networks is one of the most challenging and interesting problems in the field. In this study, we investigate the topic focusing on the interpretability of deep learning-based registration methods. In particular, with the appropriate model architecture and using a simple linear projection, we decompose the encoding space, generating a new basis, and we empirically show that this basis captures various decomposed anatomically aware geometrical transformations. We perform experiments using two different datasets focusing on lungs and hippocampus MRI. We show that such an approach can decompose the highly convoluted latent spaces of registration pipelines in an orthogonal space with several interesting properties. We hope that this work could shed some light on a better understanding of deep learning-based registration methods.

\end{abstract}

\section{Introduction}
Deep learning methods provide the state of the art performance for various applications currently. This is due to their inherent property to generate highly abstract representations hierarchically. These representations are building on top of each other, making it possible to encode highly non-linear manifolds. Even though such hierarchies can outperform traditional methods, they lack explainability, making their translation difficult to solve real-life problems. This drawback is of great significance in the medical field and especially for the algorithms that are intended to be adapted to clinical practice, addressing problems of precision medicine~\cite{holzinger2019causability,castro2020causality}. For these reasons, it is essential to identify ways to understand better the high throughput operations that are applied.

Recently, with the introduction of the differentiable spatial transformer~\cite{jaderberg2016spatial}, trainable deep learning registration methods are becoming more and more popular, reducing computational times while reporting similar to traditional methods performance~\cite{balakrishnan2019voxelmorph,stergios2018linear,mok2020fast}. Meanwhile, the deformation field, which is one of the products of deformable registration methods, has been shown to encode not only the spatial correspondences but also clinical relevant information that could add valuable aspects to a variety of problems related to survival assessment or anomaly detection~\cite{ou2015deformable}. Indeed, encoding information between subjects can be very informative for various medical tasks such as medical image segmentation~\cite{estienne2019uresnet}. However, according to our knowledge, there are not many efforts focusing on understanding and analysing this encoding information which could initiate the explainability of deep learning-based registration methods.

In this study, we propose a framework for interpreting the encoded representation of deep learning-based registration methods. In particular, with the appropriate model architecture and by using a simple linear projection, we decompose the encoding space, generating a new basis that captures various geometrical operations. This decomposed encoding space is then driving the generation of the deformation field. The contributions of this work are twofold: \textit{(i)} to the best of our knowledge, this study is one of the first to explore the explainability of deep learning-based registration methods through their encodings using linear projections, \textit{(ii)} we show empirically, using two different datasets, one focusing on lungs and the other on the brain hippocampus that our projections are associated with different types of deformations and in particular rigid transformations. We hope that this work can highlight the very challenging topic of explainability of deep neural networks.

\section{Related Work}

Explaining how deep neural networks function is a matter of extensive research the recent years. GradCam~\cite{selvaraju2017grad} is one of the most popular methods that can provide some insights on deep neural networks for many applications, including medical imaging. GradCam highlights the region of the original input that contributes the most to the final prediction, producing coarse heatmaps based on the gradients. Similar to GradCam, there are many additional methods based on the gradient~\cite{zhou2016learning,chattopadhay2018grad,springenberg2014striving} that are commonly used for the explainability of the models. Moreover, in~\cite{fong2017interpretable} the authors proposed a general framework of explanations as meta-predictors while they also reinterpret the network's saliency providing a natural generalisation of the gradient-based saliency techniques. Even though such approaches can provide information on where the models attend, they can be mostly utilised in classification or detection schemes.

Representation disentangling methodologies is a concurrent field of research also investigating explainability topics. Such approaches are mainly focusing on generating interpretable latent representations by enforcing several constraints. This can be achieved either using architecture tricks~\cite{shu2018deforming,sahasrabudhe2019lifting} or with appropriate loss functions~\cite{chen2016infogan,kingma2013auto}. In medical image computing, several studies focus on approaches for generating disentangled or decomposed representations. In~\cite{qin2019unsupervised} for example, the authors proposed a multimodal image registration method by decomposing the volumes into a common latent shape space and separate latent appearance spaces via image-to-image translation approach and generative models. 
Our method shares many common points with the approaches mentioned above, yet it focuses on exploring the registration latent space decomposition.

\section{Methodology}
Deep learning-based registration methods have received much attention in the last few years~\cite{balakrishnan2019voxelmorph,stergios2018linear}. 
Formally, let us consider two volumes, the moving $M$ and the fixed $F$. The goal of deep learning-based registration methods is to obtain the best parameters $\theta^*$ for the network $g_\theta$ that will map most accurately $M$ to $F$ using the predicted deformation grid $\Phi$. The network $g_\theta$ usually is composed of an encoding $E_\psi$ and a decoding $D_{\omega}$ part. 


\begin{figure}[t!]
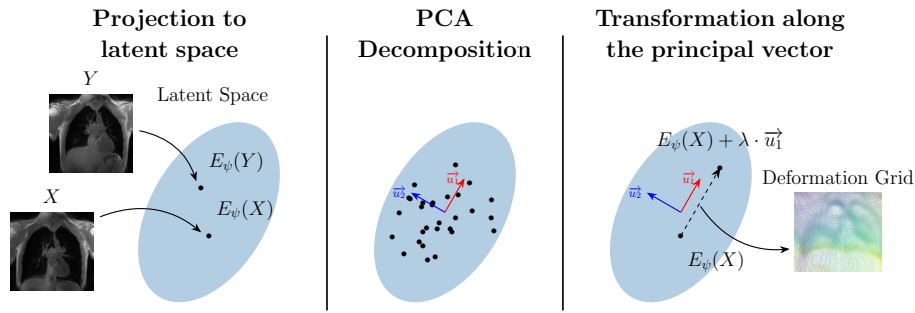

\centering
\includestandalone[width=\textwidth]{tikz_schema}
\caption{Overall overview of our proposed framework. The different subjects ($X$,$Y$) are projected on the latent representation by the encoder $E_{\psi}$ and then a linear decomposition of this latent space is calculated to identify a new vector space $(\vec{u}_{i})$.}
\label{fig:overall}
\end{figure}

There are multiple ways to fuse the input volumes in deep learning-based registration approaches. Most of the methods use an early fusion strategy on which the two volumes are concatenated before they pass through the $g_\theta$. However, some methods investigate late fusion strategies~\cite{heinrich2019closing,estienne2020deepA} where the two volumes pass independently through the encoder, and their merging operation is achieved in the encoding representation using various operations such as concatenation or subtraction. Thanks to this formulation, each volume has a unique encoding representation. In this study, we adopt the second strategy using the subtraction operation to encode each volume independently and calculate its latent space's linear decomposition. In Figure~\ref{fig:overall}, the overall scheme is presented. 

\subsection{Deep learning-based registration scheme}
To perform our experiments and obtain our embeddings, we defined a network based on a 3D UNet architecture~\cite{cicek20163d}. The encoder and the decoder are composed of a fixed number of blocks with 3D convolution layers (stride $3$, padding $1$), instance normalisation layer and leaky ReLU activation function. The down and up-sampling operations are performed with a 3D convolution layer with stride and padding of $2$. One of the main differences in our architecture was the absence of skip connections. Indeed, we want to enforce that all information passes through the last encoding layer without any leak due to the skip connections. This modification led us to reduce the downsampling operations from four to three for the lung dataset, to maintain the spatial resolution of the bottleneck.

Different formulations have been proposed to generate the deformation from deep learning schemes, 
such as displacement field formulation \cite{balakrishnan2019voxelmorph}, diffeomorphic formulations~\cite{dalca2018unsupervised,krebs2019learning} and formulations based on the spatial gradients~\cite{stergios2018linear}. In this work, we focused on the last one, with our network regressing the spatial gradients $\nabla_x \Phi_x$, $\nabla_y \Phi_y$ and $\nabla_z \Phi_z$, while the final deformation field is obtained through a cumulative sum operation. We also followed the symmetric formulation proposed in~\cite{estienne2020deepB}, predicting both the forward and backward deformations: $\nabla \Phi_{M\rightarrow F} = D_{\omega} ( E_\psi(M) - E_\psi(F) )$  and $\nabla \Phi_{F\rightarrow M} = D_{\omega} ( E_\psi(M) - E_\psi(F) )$.

The network was trained with a combination of four losses, one focusing on the intensity similarity using normalised cross-correlation ($\Lreg$), one focusing on anatomical structures using dice loss ($\Lseg$) and two losses for regularisation of the displacements. The first one was the Jacobian loss which is exploited on different works such as~\cite{mok2020fast,kuang2019faim,zhang2020diffeomorphic} ($\Ljac$) and the second one enforcing smooth gradients similar to~\cite{estienne2020deepB} ($\Lsmooth$). 
As such our final loss is:  $\mathcal{L} = (\Lreg + \Lseg + \alpha \Lsmooth + \beta \Ljac)_{\small{M\rightarrow F}} + (\Lreg +  \Lseg + \alpha  \Lsmooth + \beta  \Ljac)_{F\rightarrow M}$ with $\alpha$ and $\beta$ being the weights of the regularisation losses.



\subsection{Decomposition of latent space}
Let $A_{train} = \{X_i | i \in \left[ 0, n \right] \}$ be the set of our $n$ training samples. The proposed formulation apply the encoder independently to each volume, and thus we can obtain the set of latent vectors: $E_\psi (A_{train}) = \{ E_\psi (X_i) | i \in \left[ 0, n \right] \}$. Then, we decompose this space using principal components analysis (PCA). That way, we obtain a set of principal vectors $\mathcal{U}_{K} = ( \overrightarrow{u_1},  \cdots , \overrightarrow{u_K})$ with $K$ being a hyperparameter fixing the number of principal components. It worth noting that each vector $\vec{u}_{i}$ has the same size as the activation map of the encoder's last layer. This size depends on the number of channels, the size of the input images and the number of downsampling operations. We flatten each encoding representation from its four dimensions representation (channel dimension and the three spatial dimensions) to a one-dimensional array to perform the PCA. Thus, the PCA is not calculated channel-wise, but all the channels are considered together. Each principal vector $\vec{u}_{i}$ can be converted to a deformation grid $\phi_i$ using the corresponding decoder $D_{\omega}$: $\phi_i = D_{\omega}(\vec{u}_i)$. Therefore, we obtained a set of elementary transformations $\{ \phi_i \}_{i=1\cdots K}$. These elementary transformations generate a basis that can be used to approximate and decompose every new deformation. Using such a decomposition, we can obtain a representation in small dimensions of every training volume $X_i$. These representations are obtained by the projection of $E_\psi(X_i)$ to each principal vector: $a^j_i = E_\psi(X_i) \cdot \overrightarrow{u_j}$. For every volume of our training set we have the approximation: $E_\psi(X_i) \approx \sum_{j=1}^K a^j_i \overrightarrow{u_j}$. After calculating the vector of the principal components $\mathcal{U}_K$ with the training set, we projected each image of the validation set to obtain its PCA representation.

\subsection{Implementation and Training Details}
The Adam optimiser was used for our training, with a constant learning rate set to $1e^{-4}$, a batch size equal to $4$ and $8$ for lung and hippocampus, respectively. Our models were trained for $600$ epochs, and it last approximately $4$ and $9$ hours for the lung and hippocampus dataset. Concerning data augmentation, we applied random flip, rotation, translation and zoom.  Moreover, the weights of the different loss components were set to $1$ except the loss for smoothness set to $\alpha = 0.1$ for both datasets and the weight for the jacobian loss $\beta$ that was discarded for the hippocampus dataset. During the training process, we registered random pairs of different patients. Our training has been performed using the framework PyTorch and one GPU card Nvidia Tesla V100 with 32G memory. The PCA decomposition was calculated using the library scikit-learn, and the number of principal components $K$ was set to $32$. Using $32$ components, our decomposition covered $95\%$  and $93\%$ of the variance ratio for the lung and hippocampus dataset, respectively, while $42\%$ and $62\%$ are covered by the first four components for each dataset, respectively.

\section{Experiments and Results}
We performed our experiments on two different datasets, one public and one private. Starting with the public dataset, we conduct experiments with the hippocampus\footnote{\url{http://medicaldecathlon.com/}} \cite{simpson2019large}. This dataset comprises $394$ MRI with the segmentations of two small structures, the head and the body hippocampus. The images have been cropped around the hippocampus into small patches of $64\times 64\times 64$ voxels. The second dataset is composed of $41$ lung MRI patients ($12$ healthy and $29$ diseased with pulmonary fibrosis) together with their lung segmentations. Each patient had been acquired in two states, the inspiration and the expiration. Each volume has been resampled to $1.39$mm on the x and z-axis and $1.69$ on the y-axis and cropped to $128\times 64\times 128$ volumes. The same normalisation strategy has been applied for the two datasets:  $\mathcal{N}(0,1)$ standardisation, clip to $\left[-5,5\right]$ to remove outliers values and min-max normalisation to $(0,1)$. Both datasets were split into training and validation, resulting in $200$ and $60$ patients for hippocampus and $28$ and $13$ patients for the lung dataset.

As the first step of our evaluation, we benchmarked the performance of the registration network $g_\theta$, on which our decomposition is based on. More specifically, we obtained a Dice coefficient of $0.90\pm0.04$ for the lungs and $0.76\pm0.05$ for the hippocampus, while the initial unregistered cases reported a Dice of $0.74\pm0.14$ and $0.59\pm0.15$ respectively. Moreover, we calculated the registration for $g_\theta$ with the skip connections to measure their impact on the registration. The Dice is then equal to $0.92\pm0.02$ and $0.85\pm0.03$ respectively. Thus, by removing the skip-connections, we decrease the performance of the registration, slightly on the lungs, more importantly, on the hippocampus. However, both strategies register the pair of volumes properly.

\subsection{Qualitative Evaluation}
To understand and evaluate the calculated components of $\mathcal{U}_{K}$ per dataset, we perform a qualitative analysis. In particular, for each principal vector $\overrightarrow{u_i}$, we calculated the corresponding deformation $\{ \phi_i \}_{i=1\cdots K}$ and we applied to the moving image $M$ together with its corresponding segmentation map $M_{seg}$. More formally, the deformed contour correspond to $\mathcal{W}(D_{\omega}(\lambda \overrightarrow{u_i}), M_{seg})$ with $\mathcal{W}$ being the warping operation and $\lambda$ the parameter to control the strength of the displacements for better visualisation.

\begin{figure}[t!]
  \centering
  
  \begin{minipage}[t]{.25\textwidth}%
  \centering
  \includegraphics[height=\height,trim=2.5cm 0.8cm 2.5cm 0.8cm,clip]{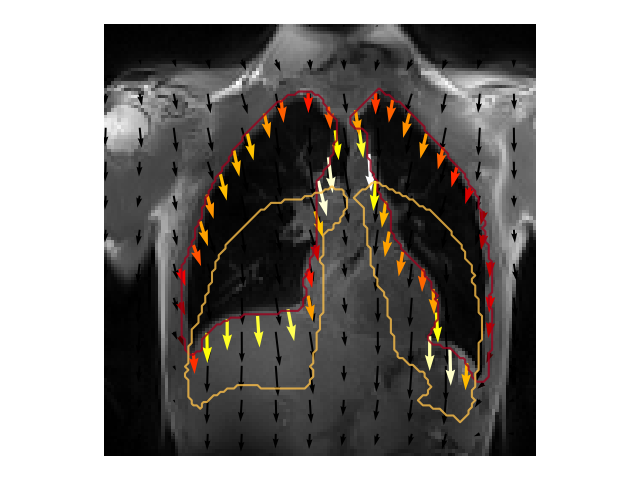}
  \end{minipage}%
  \begin{minipage}[t]{.25\textwidth}%
  \centering
  \includegraphics[height=\height,trim=5.3cm 0.8cm 5.3cm 0.8cm,clip]{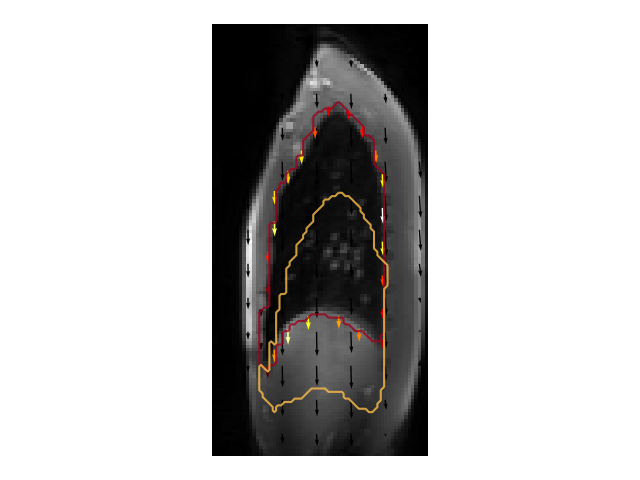}
  \end{minipage}%
  \begin{minipage}[t]{.25\textwidth}%
  \centering
  \includegraphics[height=\height,trim=2.5cm 0.8cm 2.5cm 0.8cm,clip]{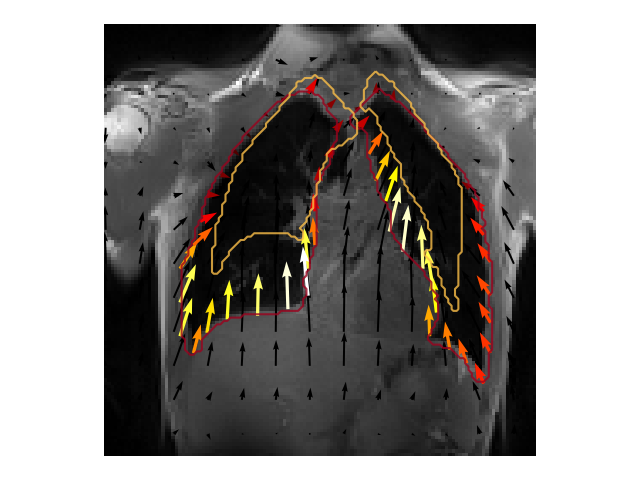}
  \end{minipage}%
  \begin{minipage}[t]{.25\textwidth}%
  \centering
  \includegraphics[height=\height,trim=5.3cm 0.8cm 5.3cm 0.8cm,clip]{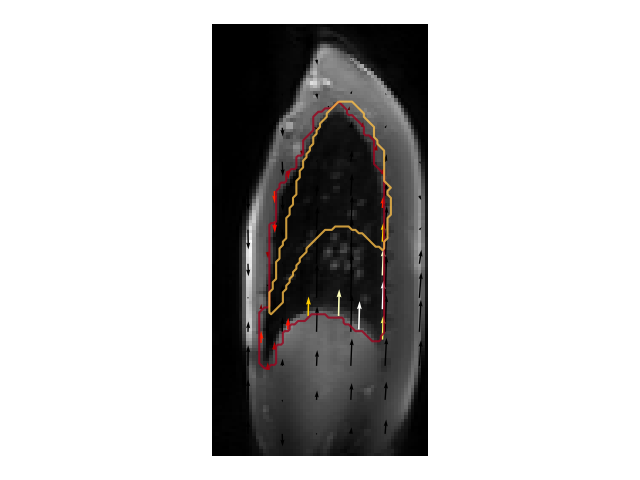}
  \end{minipage}%
  
  \begin{minipage}[t]{.25\textwidth}%
  \centering
  \includegraphics[height=\height,trim=2.5cm 0.8cm 2.5cm 0.8cm,clip]{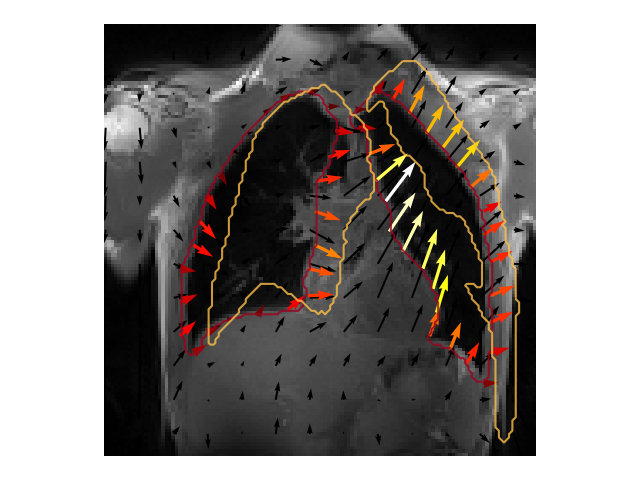}
  \end{minipage}%
  \begin{minipage}[t]{.25\textwidth}%
  \centering
  \includegraphics[height=\height,trim=5.3cm 0.8cm 5.3cm 0.8cm,clip]{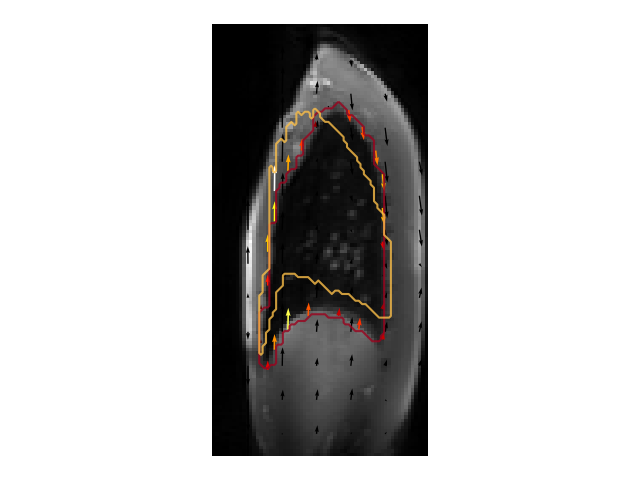}
  \end{minipage}%
  \begin{minipage}[t]{.25\textwidth}%
  \centering
  \includegraphics[height=\height,trim=2.5cm 0.8cm 2.5cm 0.8cm,clip]{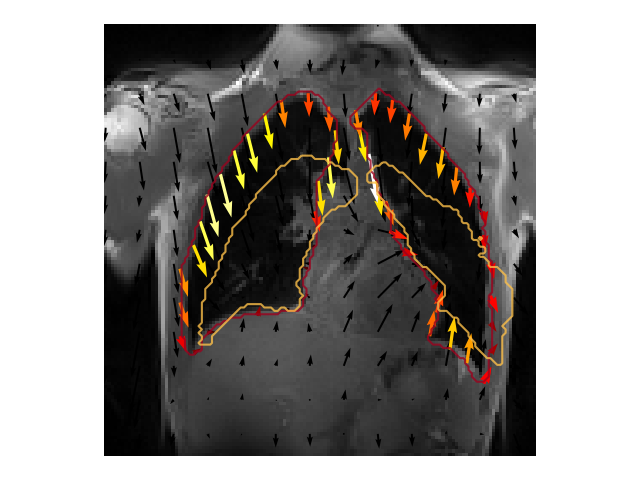}
  \end{minipage}%
  \begin{minipage}[t]{.25\textwidth}%
  \centering
  \includegraphics[height=\height,trim=5.3cm 0.8cm 5.3cm 0.8cm,clip]{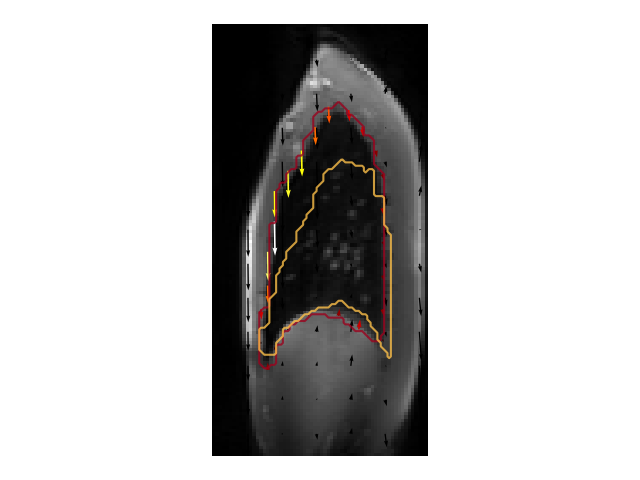}
  \end{minipage}%

  \caption{Visualisation of the displacements following the first four principal components. For each component, we depicted coronal and sagittal views. In red, the contours of the lungs of the $M$, and in gold, the $\mathcal{W}(D_{\omega}(\lambda \protect\overrightarrow{u_i}),M)$ lung's contours. The deformation field is represented with arrows. The arrows' norm is represented with a colour map, red being the smallest and white the largest. Other patients and components are displayed on supplementary materials.}
  \label{fig:lung_pca}
\end{figure}

\begin{figure} [t!]
    \centering
        \begin{minipage}[t]{.2\textwidth}%
        \centering
        \includegraphics[width=\linewidth,trim=3.2cm 1cm 3.2cm 1.5cm,clip]{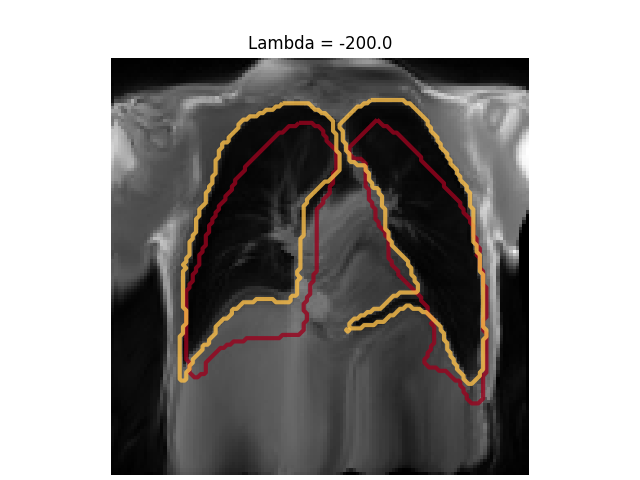}
        \end{minipage}%
        \begin{minipage}[t]{.2\textwidth}%
        \centering
        \includegraphics[width=\linewidth,trim=3.2cm 1cm 3.2cm 1.5cm,clip]{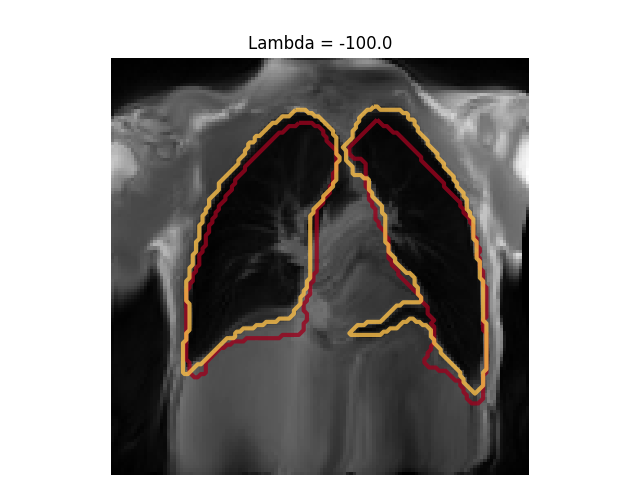}
        \end{minipage}%
        \begin{minipage}[t]{.2\textwidth}%
        \centering
        \includegraphics[width=\linewidth,trim=3.2cm 1cm 3.2cm 1.5cm,clip]{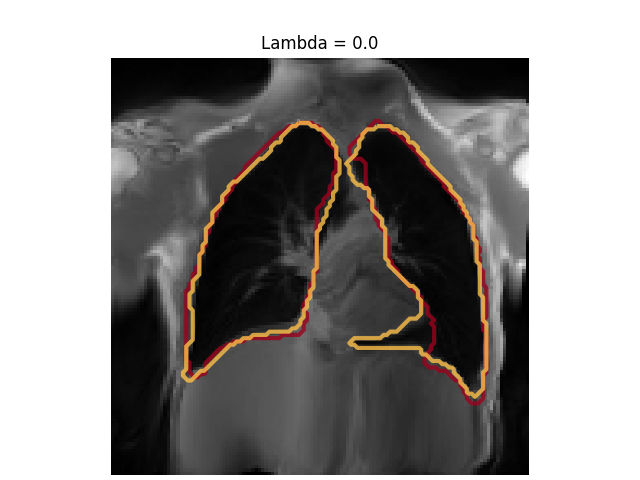}
        \end{minipage}%
        \begin{minipage}[t]{.2\textwidth}%
        \centering
        \includegraphics[width=\linewidth,trim=3.2cm 1cm 3.2cm 1.5cm,clip]{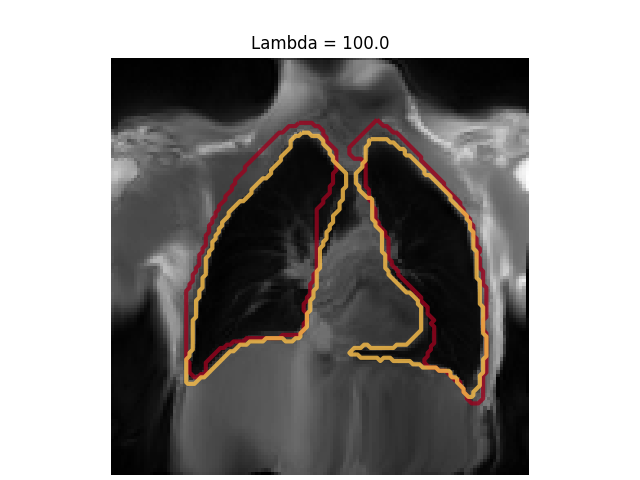}
        \end{minipage}%
        \begin{minipage}[t]{.2\textwidth}%
        \centering
        \includegraphics[width=\linewidth,trim=3.2cm 1cm 3.2cm 1.5cm,clip]{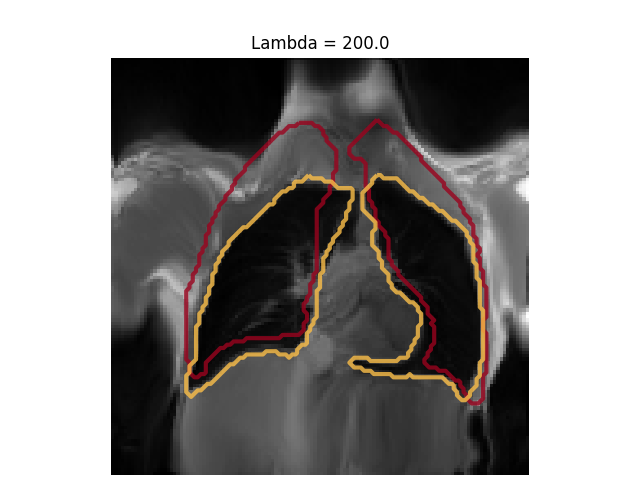}
        \end{minipage}%
        
        \begin{minipage}[t]{.20\textwidth}%
        \centering
        \includegraphics[width=\linewidth,trim=3.2cm 1cm 3.2cm 1.5cm,clip]{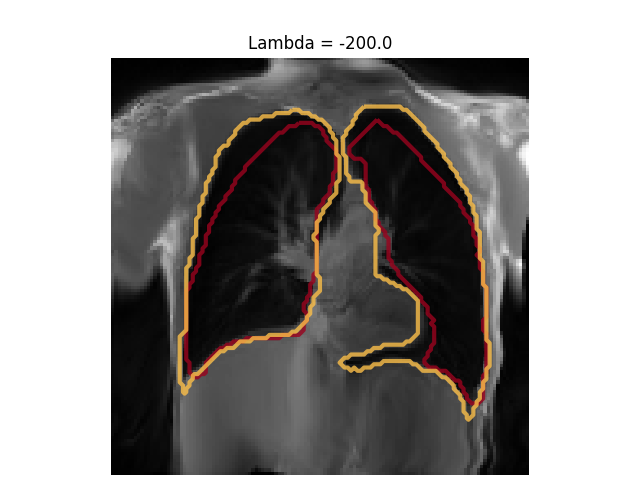}
        \end{minipage}%
        \begin{minipage}[t]{.20\textwidth}%
        \centering
        \includegraphics[width=\linewidth,trim=3.2cm 1cm 3.2cm 1.5cm,clip]{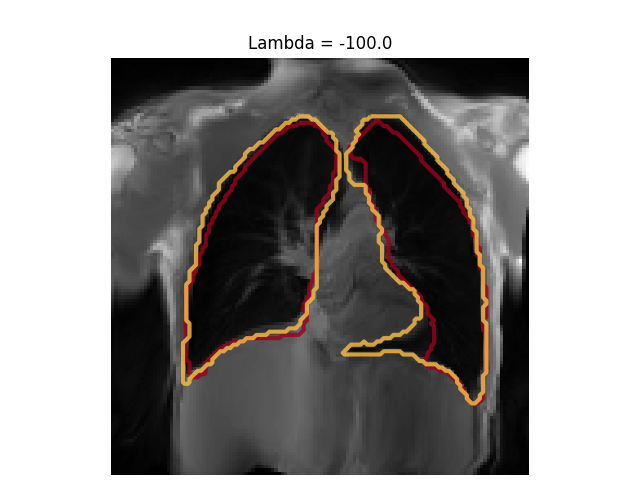}
        \end{minipage}%
        \begin{minipage}[t]{.20\textwidth}%
        \centering
        \includegraphics[width=\linewidth,trim=3.2cm 1cm 3.2cm 1.5cm,clip]{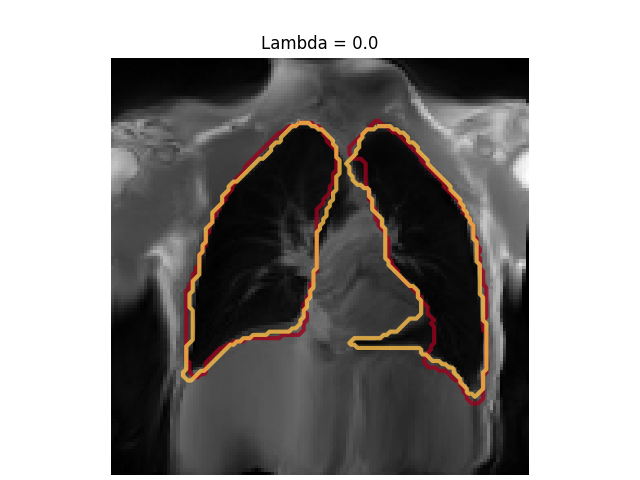}
        \end{minipage}%
        \begin{minipage}[t]{.20\textwidth}%
        \centering
        \includegraphics[width=\linewidth,trim=3.2cm 1cm 3.2cm 1.5cm,clip]{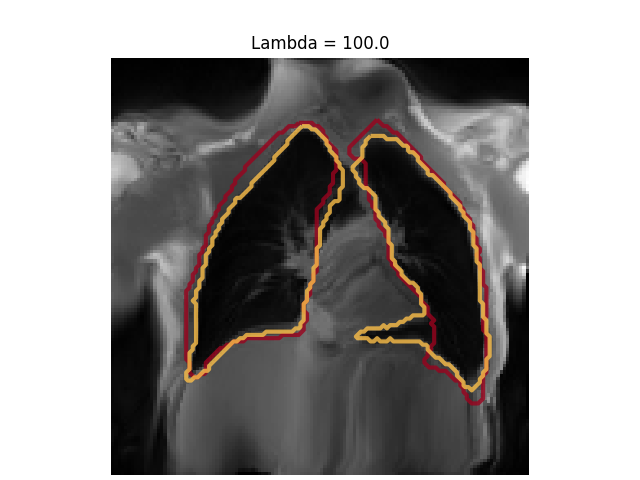}
        \end{minipage}%
        \begin{minipage}[t]{.20\textwidth}%
        \centering
        \includegraphics[width=\linewidth,trim=3.2cm 1cm 3.2cm 1.5cm,clip]{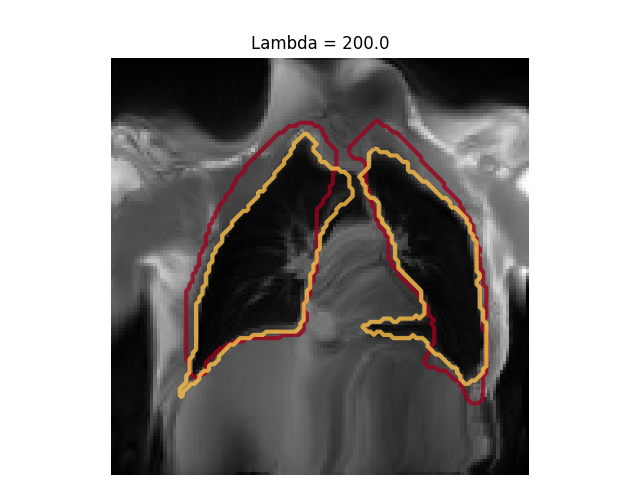}
        \end{minipage}%
    \caption{Representation of the deformed MR together with its lung contours following the first and fourth principal components $u_1$, $u_4$. The red contour represents the position of the lung segmentation of the input image while the gold contour the position of the deformed lung. The values of lambda range from: $-200$, $-100$, $0$, $100$ and $200$ (left to right). Negative values of lambda correspond to an upward translation, while positive values to a downward translation.}
    \label{fig:lambda}
\end{figure}

In Figure~\ref{fig:lung_pca}, we show the principal components obtained for one validation subject for the lung dataset. Interestingly, one can observe that each $\phi_i$ corresponds to a different elementary transformation. More precisely, the $1^{st}$ component is associated with translation, the $2^{nd}$ with a deformation focusing on the bottom of the lungs, the $3^{rd}$ with a deformation on the right lung focusing also on the heart region and lastly the $4^{th}$ with a deformation focusing on the top region of the lung and shoulders. 
\begin{figure}[b!]
  \centering
  
  \begin{minipage}[t]{.25\textwidth}%
  \centering
  \includegraphics[width=\linewidth,trim=4cm 3cm 4cm 2.8cm,clip]{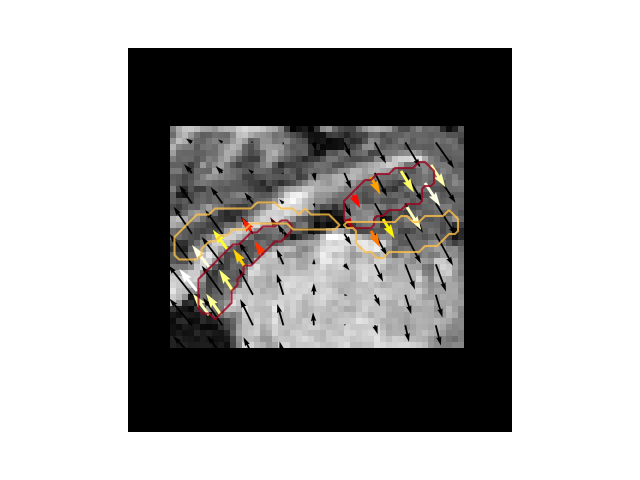}
  \end{minipage}%
  \begin{minipage}[t]{.25\textwidth}%
  \centering
  \includegraphics[width=\linewidth,trim=4cm 3cm 4cm 2.8cm,clip]{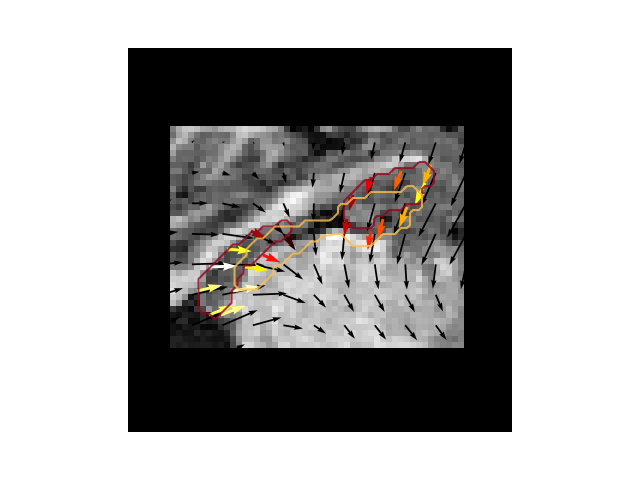}
  \end{minipage}%
  \begin{minipage}[t]{.25\textwidth}%
  \centering
  \includegraphics[width=\linewidth,trim=4cm 3cm 4cm 2.8cm,clip]{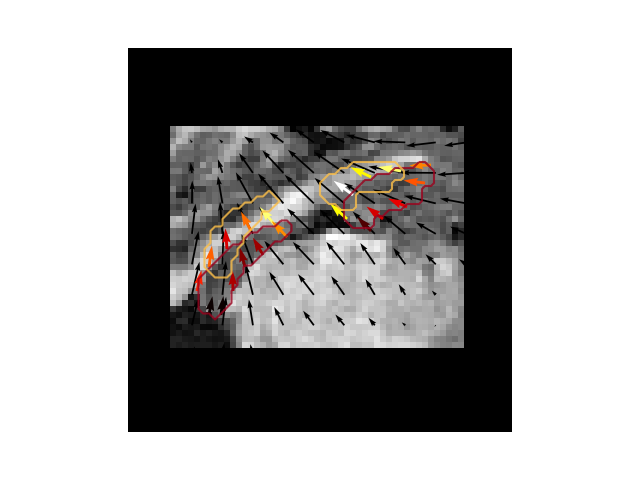}
  \end{minipage}%
  \begin{minipage}[t]{.25\textwidth}%
  \centering
  \includegraphics[width=\linewidth,trim=4cm 3cm 4cm 2.8cm,clip]{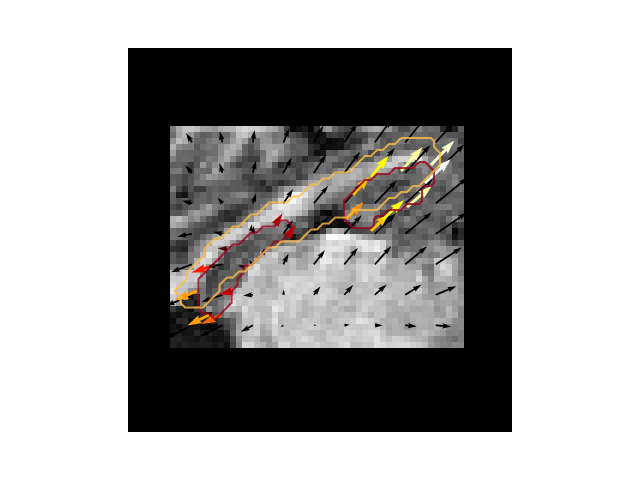}
  \end{minipage}%
  
  \caption{Visualisation of the displacements following the first four principal components. We depicted a sagittal view of one patient of the validation set of the hippocampus dataset. We represented the ground truth hippocampus contours (red) and the deformed one (gold), following the principal components.}
  \label{fig:hipo_pca}
\end{figure}

In Figure~\ref{fig:lambda}, we show the effect of the values of $\lambda$. In the figure, we present the lung contours of the scaled component (in red) and the corresponding component of the warped of the first ad third components. As we have indicated, the $1st$ component is associated with translation, which we can also be observed in this visualisation. In particular, for this experiment we sample $\lambda$ from the values $\{ -200, -100, 0, 100, 200\}$. One can observe that we retrieve a near identity deformation for a value of $0$, while for negative and positive values, the lung moves up and down, respectively. On the other hand, the fourth component is responsible for deforming the shoulders and the top of the lungs. Indeed, one can observe that through the different $\lambda$ values, the top lungs region is the one that reports the most changes. In Figure~\ref{fig:hipo_pca}, similarly, the $4$ deformations produced by the first $4$ principal components of the hippocampus dataset are presented. In this case, the $1^{st}$ component seems to capture rotation on the sagittal plane, the $2^{nd}$ translation and shrinking towards the bottom right, while the $3^{rd}$ seems to be the same operation towards the top left corner. Finally, the $4^{th}$ seems to be related to scaling, inflating both the hippocampus's head and tail. We observed that the decomposition of the two datasets created different elementary transformations $\phi_i$, with transformations closer to affine for the hippocampus and more complex for the lung.

%

\begin{figure}[h!]
    \centering
        \begin{minipage}[t]{.33\textwidth}%
        \centering
        \includegraphics[width=\linewidth,trim=0.5cm 4.5cm 2cm 4cm,clip]{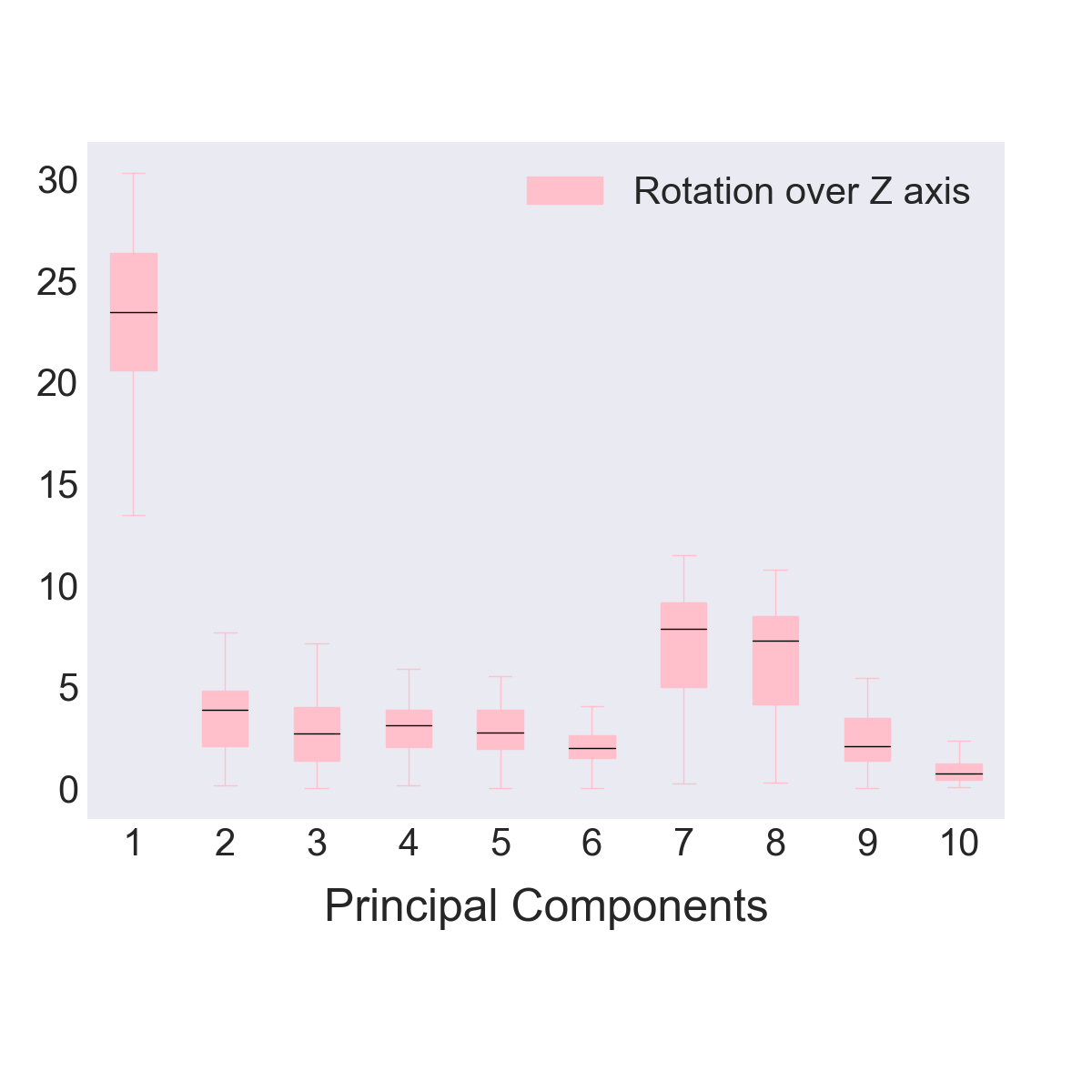}
        \end{minipage}%
        \begin{minipage}[t]{.33\textwidth}%
        \centering
        \includegraphics[width=\linewidth,trim=0.5cm 4.5cm 2cm 4cm,clip]{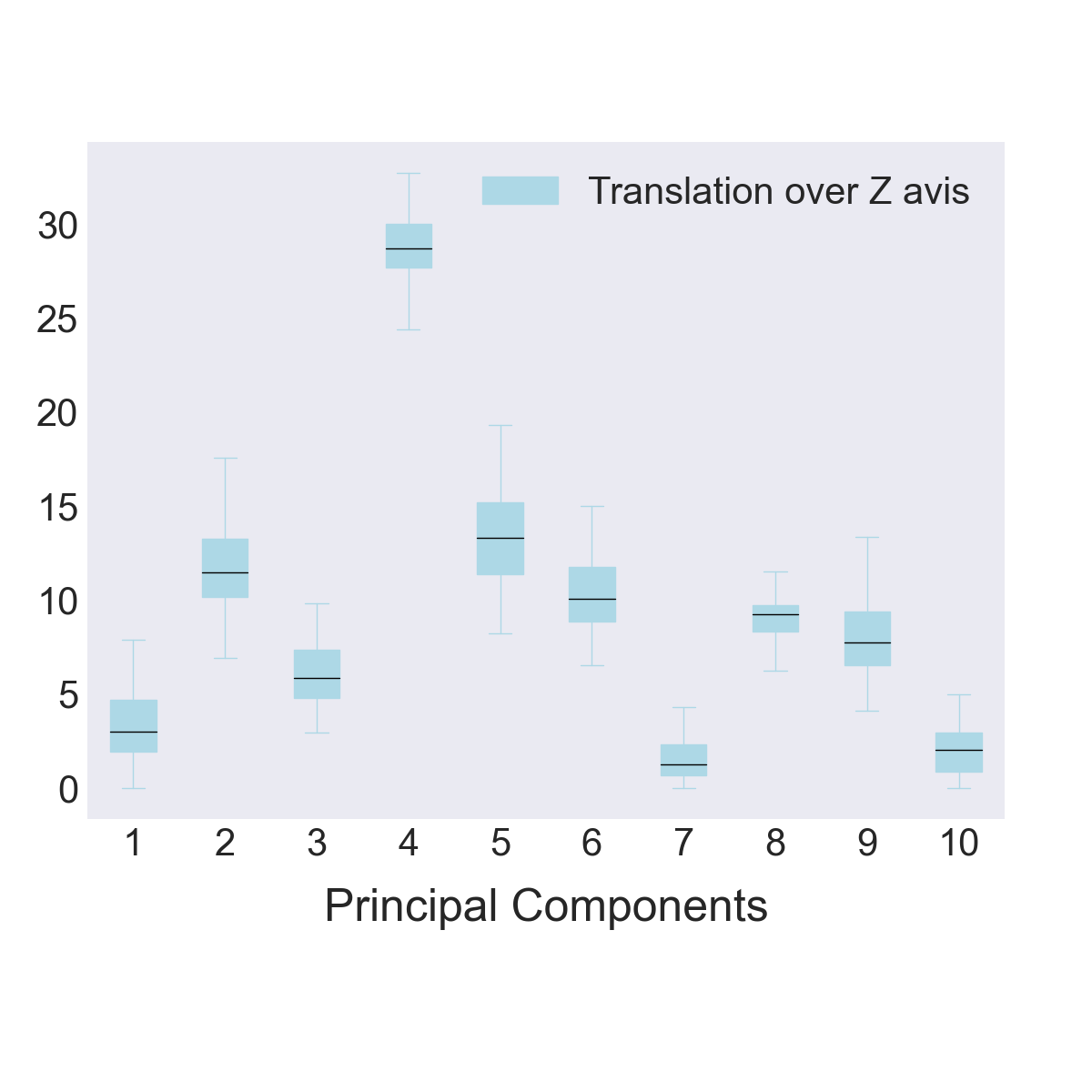}
        \end{minipage}%
        \begin{minipage}[t]{.33\textwidth}%
        \centering
        \includegraphics[width=\linewidth,trim=0.5cm 4.5cm 2cm 4cm,clip]{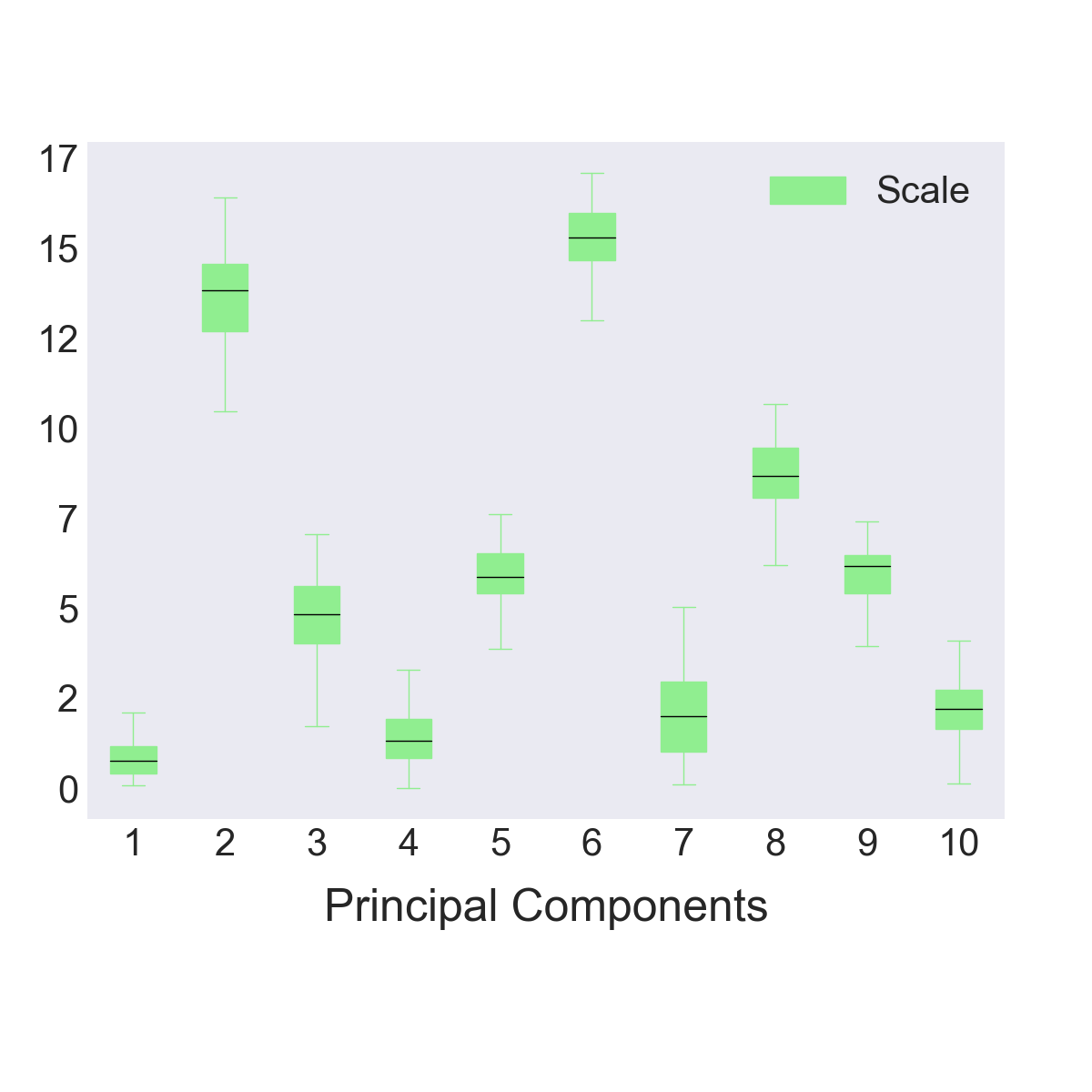}
        \end{minipage}%

    \caption{Visualisation of the differences between the components of a reference image and the same image to which we applied a predefined transformation. The first ten components have been displayed. From right to left: rotation, translation along Z-axis and scaling.}
    \label{fig:affine_transform}
\end{figure}

Finally, to verify the obtained decomposition, we performed a case study for all the validation subjects of the hippocampus dataset. More specifically, we applied some predefined translation using $10$ pixels on the $z$ axis, rotation using $20$ degrees on the $z$ axis and scaling using a factor of $0.2$, transforming each subject $X$ to $X'$. Then we calculated the difference between the projection of $E_\psi(X)$ and $E_\psi(X')$ on the PCA decomposition. In Figure~\ref{fig:affine_transform}, a box plot for all the validation subjects of the absolute difference is presented. Specifically, the amount  $||a^ j_{E_\psi(X)} - a^j_{E_\psi(X')}||$ is shown for each principal component $j$, with $a^j_{E_\psi(X)}$ being the projection of $E_\psi(X)$ on the principal vectors $\mathcal{U}_K$, for the three different applied deformations. One can observe that for rotation and translation, only one component is significantly different from the rest. In the case of scaling, however, two components seem to be more activated. Moreover, these findings are in accordance with Figure~\ref{fig:hipo_pca} for the rotation and translation. In supplementary materials, we upgraded the Figure~\ref{fig:affine_transform} by comparing the network with and without skip-connections. Contrary to our proposed formulation, many components are activated with the skip-connections, demonstrating the necessity of removing them to have a good decomposition.

\section{Discussion \& Conclusion}
In this work, we proposed an approach to decompose and explain the representations of deep learning-based registration methods. The proposed method utilises a linear decomposition on the latent space projecting it to principal components closely associated with anatomically aware deformations. Our method's dynamics are demonstrated in two different MRI datasets, focusing on lung and hippocampus anatomies. We hope that these results will take some steps towards a better understanding of latent representations learned by the deep learning registration architectures. We also explored a direct application of the PCA on the deformation's grid instead of the latent representation. However, we did not observe any qualitative correlations with types of deformations, which is the case for our proposed formulation. One of the main limitations of our approach is the difficulty of quantitative evaluation.  Our future steps include the more extensive evaluation of our method, including new anatomies such as abdominal volumes and its clinical significance. More specifically, we want to apply our approach to multi-temporal follow-up of patients, monitoring diseases' progression. 

\section*{Funding}
This work has been partially funding by the ARC: Grant SIGNIT201801286, the Fondation pour la Recherche M\'edicale: Grant DIC20161236437, SIRIC-SOCRATE 2.0, ITMO Cancer, Institut National du Cancer (INCa) and Amazon Web Services (AWS).

\newpage
\bibliographystyle{splncs04}
\bibliography{betterbibtex}

\newpage
\section*{Supplementary Material}

\begin{figure}[h!]
  \centering
  
  \begin{minipage}[t]{.25\textwidth}
  \centering
  \includegraphics[height=\height,trim=2.5cm 0.8cm 2.5cm 0.8cm,clip]{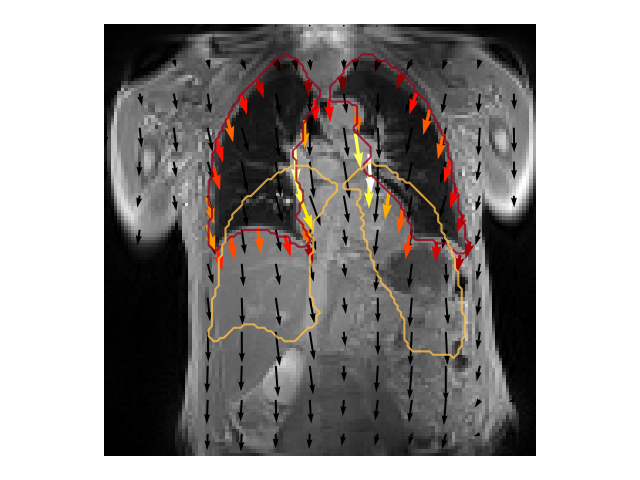}
  \end{minipage}%
  \begin{minipage}[t]{.25\textwidth}
  \centering
  \includegraphics[height=\height,trim=5.3cm 0.8cm 5.3cm 0.8cm,clip]{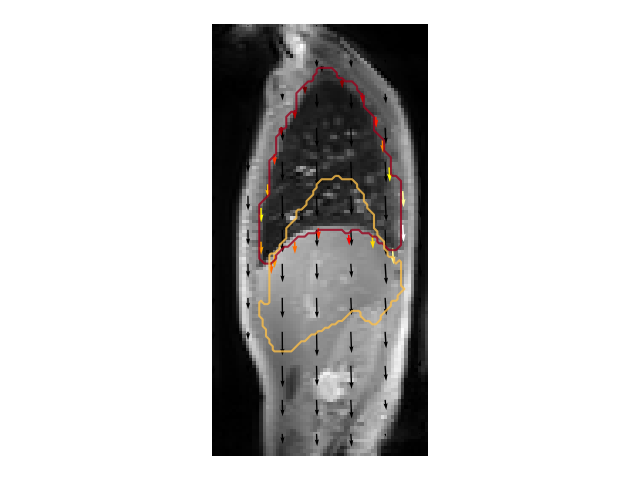}
  \end{minipage}%
  \begin{minipage}[t]{.25\textwidth}
  \centering
  \includegraphics[height=\height,trim=2.5cm 0.8cm 2.5cm 0.8cm,clip]{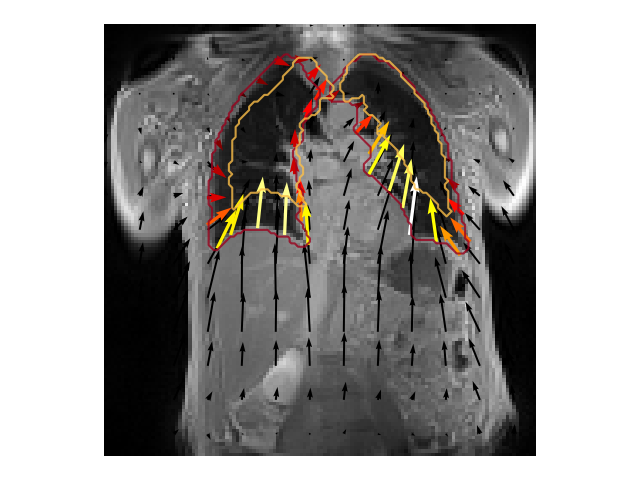}
  \end{minipage}%
  \begin{minipage}[t]{.25\textwidth}
  \centering
  \includegraphics[height=\height,trim=5.3cm 0.8cm 5.3cm 0.8cm,clip]{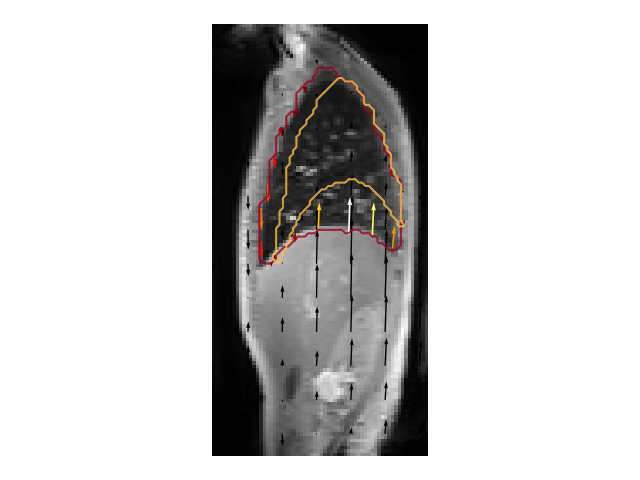}
  \end{minipage}%

  \begin{minipage}[t]{.25\textwidth}
  \centering
  \includegraphics[height=\height,trim=2.5cm 0.8cm 2.5cm 0.8cm,clip]{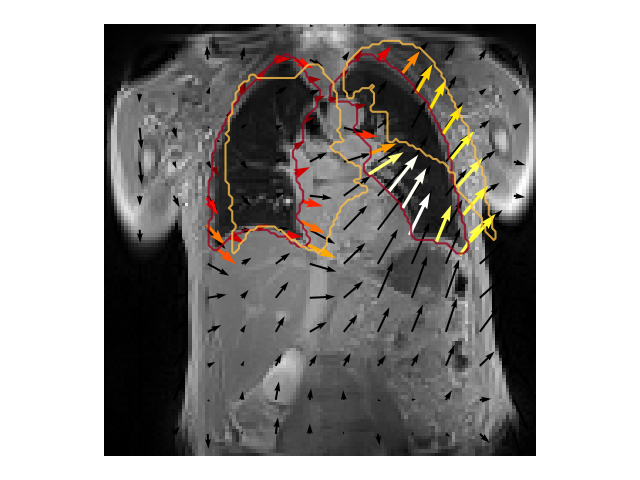}
  \end{minipage}%
  \begin{minipage}[t]{.25\textwidth}
  \centering
  \includegraphics[height=\height,trim=5.3cm 0.8cm 5.3cm 0.8cm,clip]{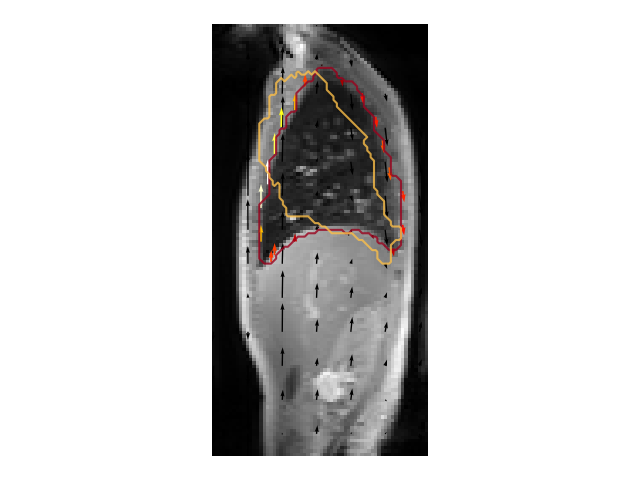}
  \end{minipage}%
  \begin{minipage}[t]{.25\textwidth}
  \centering
  \includegraphics[height=\height,trim=2.5cm 0.8cm 2.5cm 0.8cm,clip]{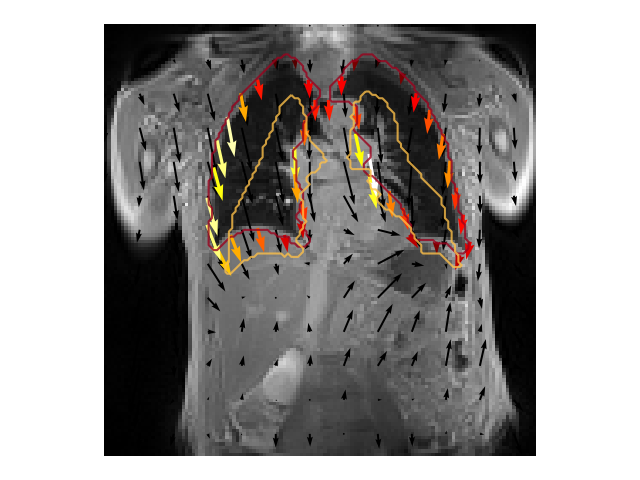}
  \end{minipage}%
  \begin{minipage}[t]{.25\textwidth}
  \centering
  \includegraphics[height=\height,trim=5.3cm 0.8cm 5.3cm 0.8cm,clip]{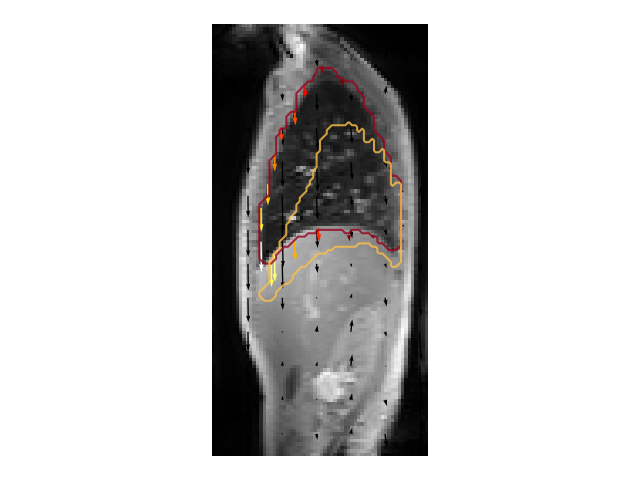}
  \end{minipage}%

\vspace{0.5cm}
  
    \begin{minipage}[t]{.25\textwidth}
  \centering
  \includegraphics[height=\height,trim=2.5cm 0.8cm 2.5cm 0.8cm,clip]{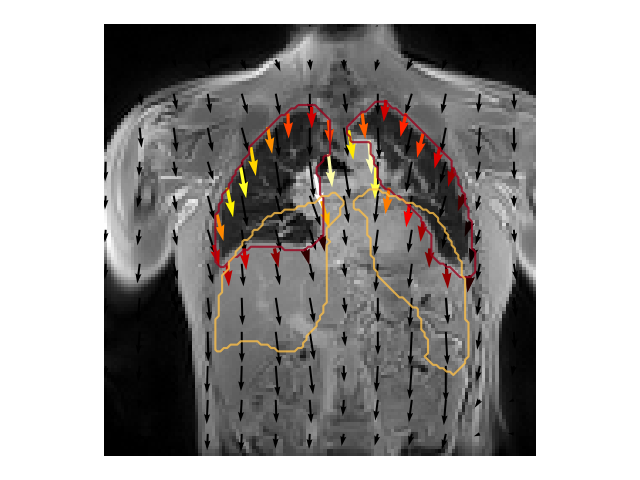}
  \end{minipage}%
  \begin{minipage}[t]{.25\textwidth}
  \centering
  \includegraphics[height=\height,trim=5.3cm 0.8cm 5.3cm 0.8cm,clip]{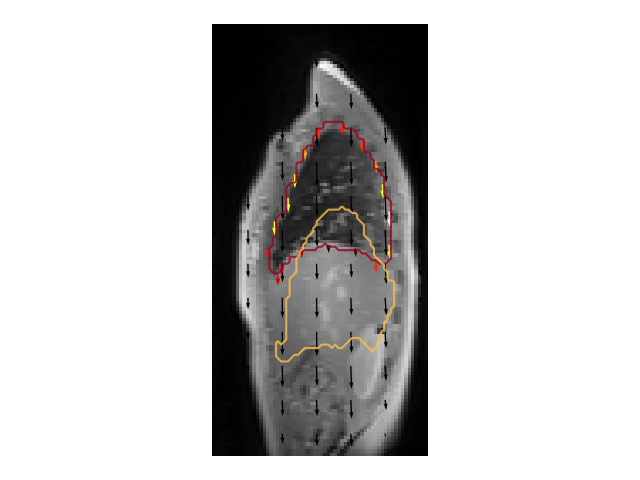}
  \end{minipage}%
  \begin{minipage}[t]{.25\textwidth}
  \centering
  \includegraphics[height=\height,trim=2.5cm 0.8cm 2.5cm 0.8cm,clip]{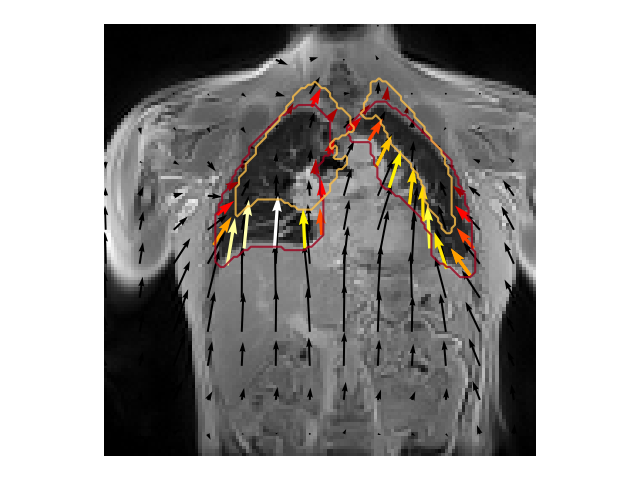}
  \end{minipage}%
  \begin{minipage}[t]{.25\textwidth}
  \centering
  \includegraphics[height=\height,trim=5.3cm 0.8cm 5.3cm 0.8cm,clip]{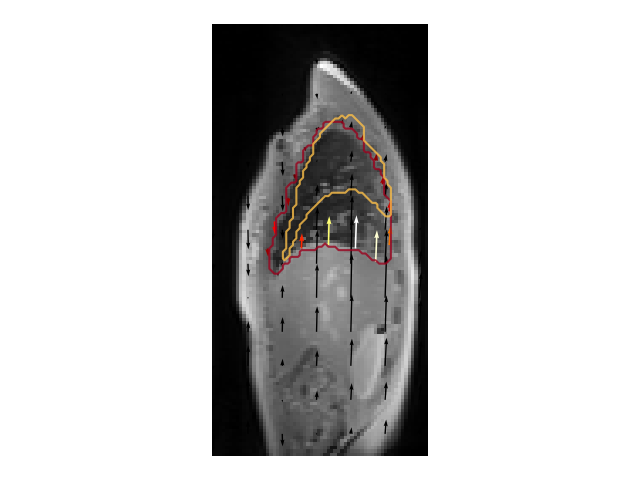}
  \end{minipage}%
  
    \begin{minipage}[t]{.25\textwidth}
  \centering
  \includegraphics[height=\height,trim=2.5cm 0.8cm 2.5cm 0.8cm,clip]{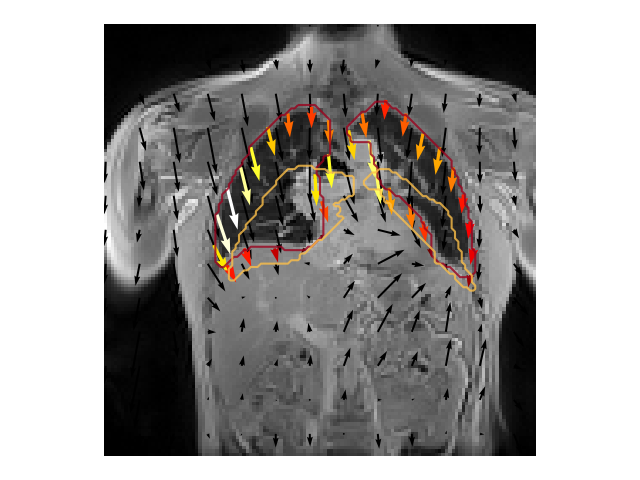}
  \end{minipage}%
  \begin{minipage}[t]{.25\textwidth}
  \centering
  \includegraphics[height=\height,trim=5.3cm 0.8cm 5.3cm 0.8cm,clip]{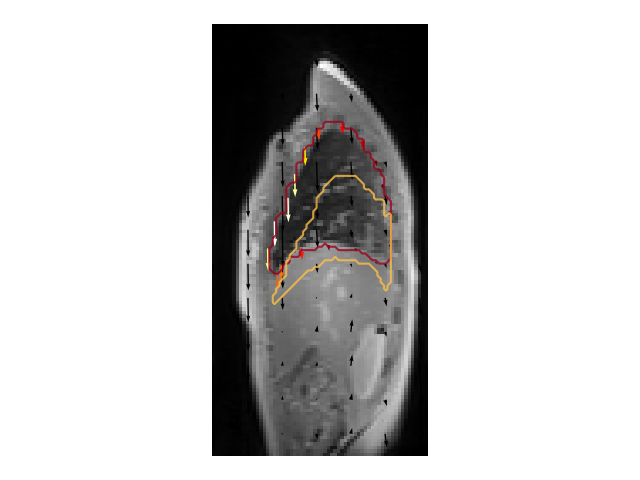}
  \end{minipage}%
  \begin{minipage}[t]{.25\textwidth}
  \centering
  \includegraphics[height=\height,trim=2.5cm 0.8cm 2.5cm 0.8cm,clip]{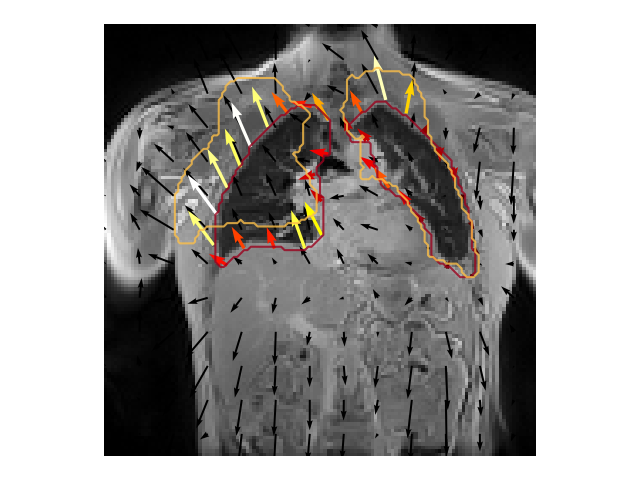}
  \end{minipage}%
  \begin{minipage}[t]{.25\textwidth}
  \centering
  \includegraphics[height=\height,trim=5.3cm 0.8cm 5.3cm 0.8cm,clip]{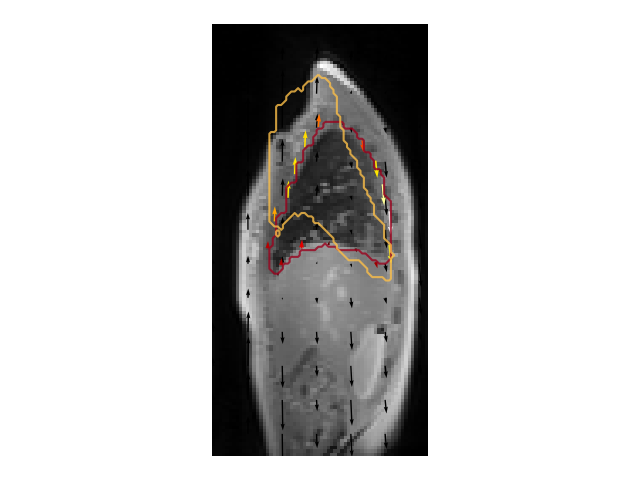}
  \end{minipage}%
  
  \caption{Extension of the figure 2 with two others patients and the component 1 to 4.}

\end{figure}

\begin{figure}[t!]
  \centering
  
  \begin{minipage}[t]{.25\textwidth}
  \centering
  \includegraphics[width=\linewidth,trim=4cm 3cm 4cm 2.8cm,clip]{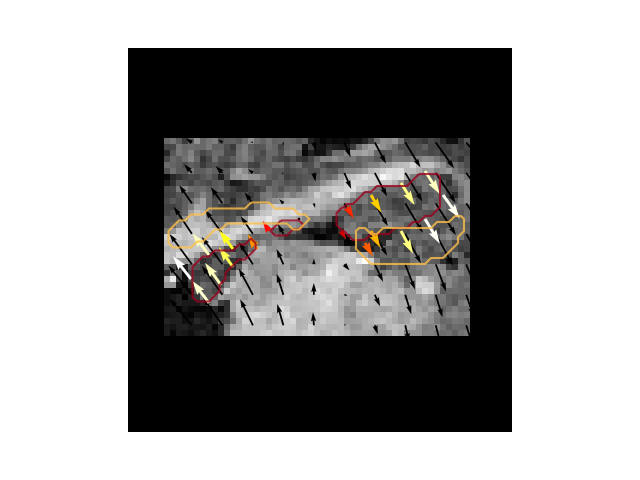}
  \end{minipage}%
  \begin{minipage}[t]{.25\textwidth}
  \centering
  \includegraphics[width=\linewidth,trim=4cm 3cm 4cm 2.8cm,clip]{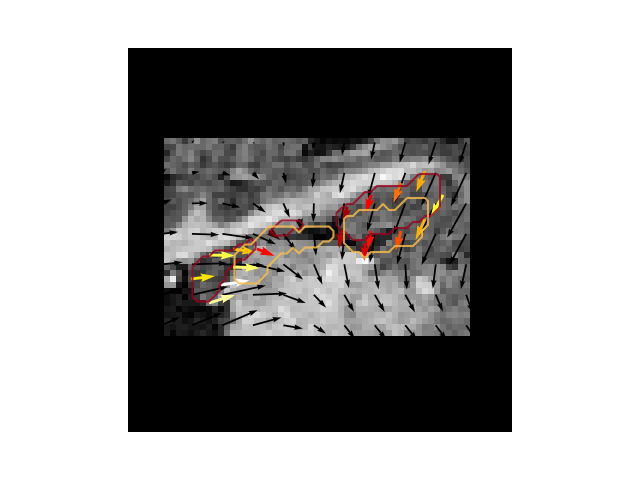}
  \end{minipage}%
  \begin{minipage}[t]{.25\textwidth}
  \centering
  \includegraphics[width=\linewidth,trim=4cm 3cm 4cm 2.8cm,clip]{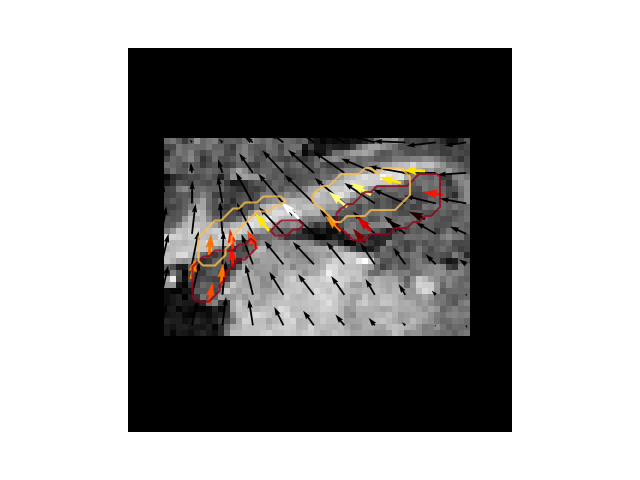}
  \end{minipage}%
  \begin{minipage}[t]{.25\textwidth}
  \centering
  \includegraphics[width=\linewidth,trim=4cm 3cm 4cm 2.8cm,clip]{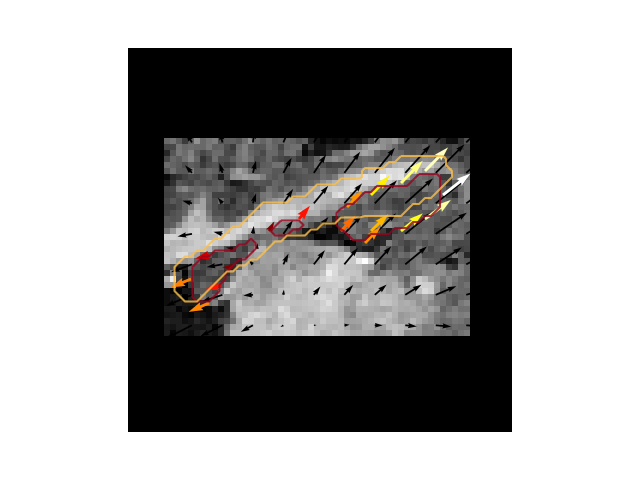}
  \end{minipage}%
  
    \begin{minipage}[t]{.25\textwidth}
  \centering
  \includegraphics[width=\linewidth,trim=4cm 3cm 4cm 2.8cm,clip]{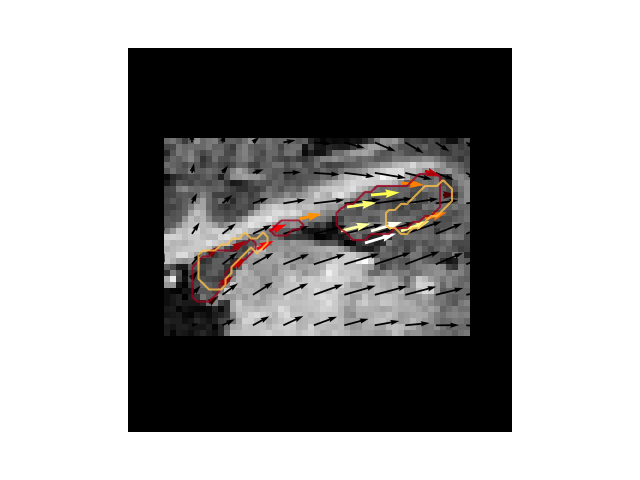}
  \end{minipage}%
  \begin{minipage}[t]{.25\textwidth}
  \centering
  \includegraphics[width=\linewidth,trim=4cm 3cm 4cm 2.8cm,clip]{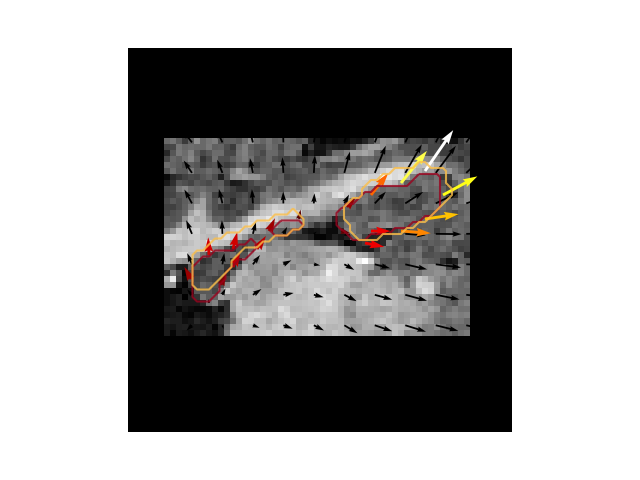}
  \end{minipage}%
  \begin{minipage}[t]{.25\textwidth}
  \centering
  \includegraphics[width=\linewidth,trim=4cm 3cm 4cm 2.8cm,clip]{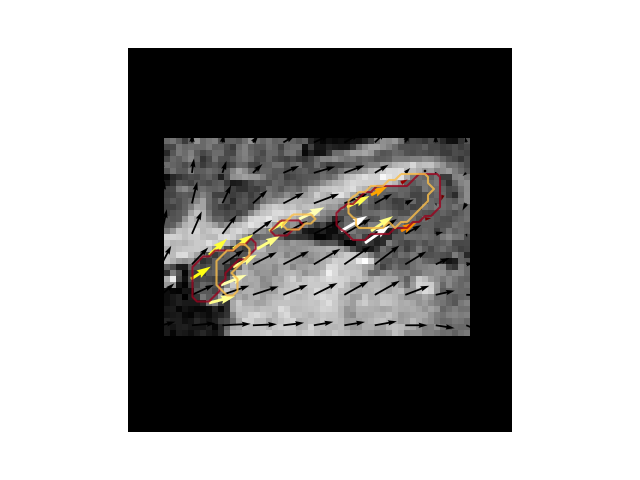}
  \end{minipage}%
  \begin{minipage}[t]{.25\textwidth}
  \centering
  \includegraphics[width=\linewidth,trim=4cm 3cm 4cm 2.8cm,clip]{Supplementary/316_axis_2_slice_32_component_6.png}
  \end{minipage}%
  
  \vspace{0.5cm}
  
    \begin{minipage}[t]{.25\textwidth}
  \centering
  \includegraphics[width=\linewidth,trim=4cm 3cm 4cm 2.8cm,clip]{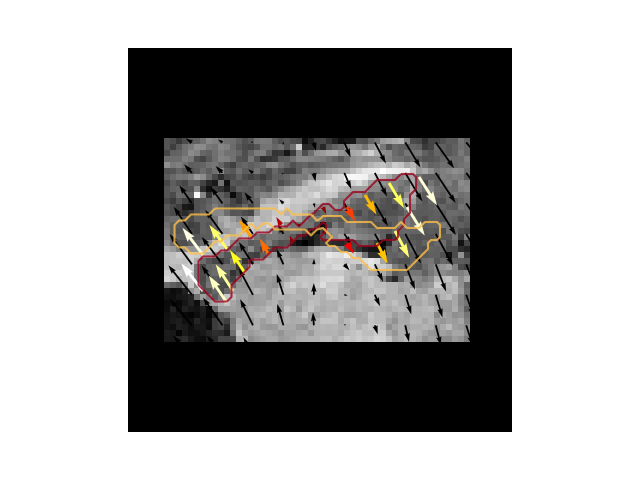}
  \end{minipage}%
  \begin{minipage}[t]{.25\textwidth}
  \centering
  \includegraphics[width=\linewidth,trim=4cm 3cm 4cm 2.8cm,clip]{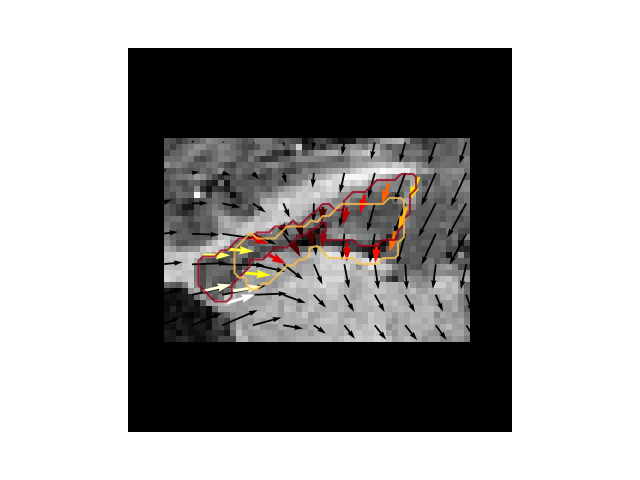}
  \end{minipage}%
  \begin{minipage}[t]{.25\textwidth}
  \centering
  \includegraphics[width=\linewidth,trim=4cm 3cm 4cm 2.8cm,clip]{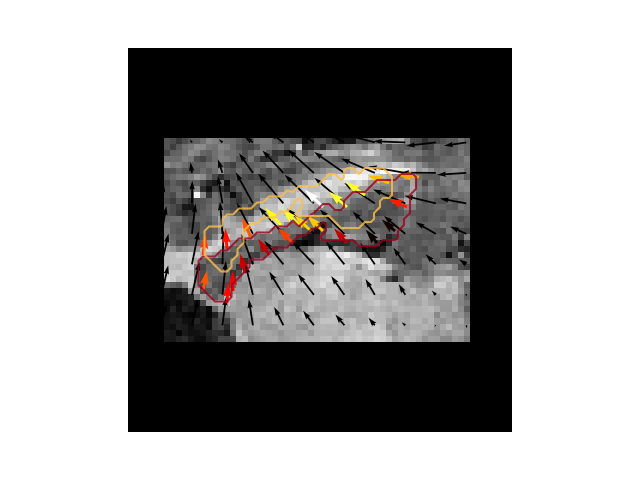}
  \end{minipage}%
  \begin{minipage}[t]{.25\textwidth}
  \centering
  \includegraphics[width=\linewidth,trim=4cm 3cm 4cm 2.8cm,clip]{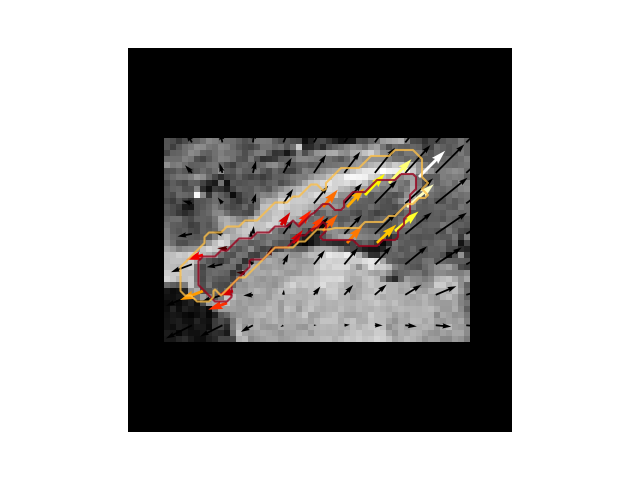}
  \end{minipage}%
  
   \begin{minipage}[t]{.25\textwidth}
  \centering
  \includegraphics[width=\linewidth,trim=4cm 3cm 4cm 2.8cm,clip]{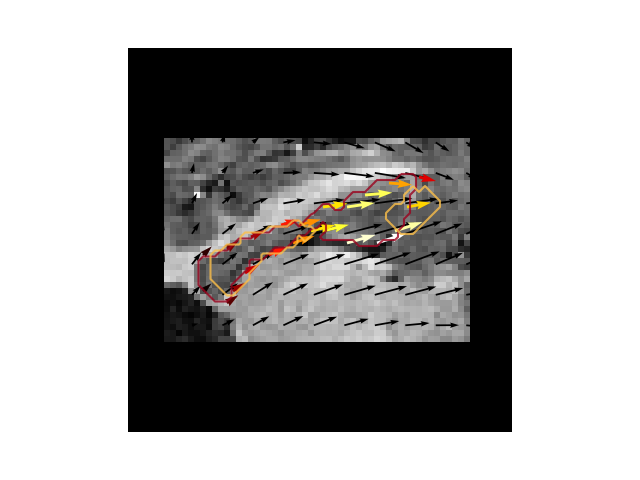}
  \end{minipage}%
  \begin{minipage}[t]{.25\textwidth}
  \centering
  \includegraphics[width=\linewidth,trim=4cm 3cm 4cm 2.8cm,clip]{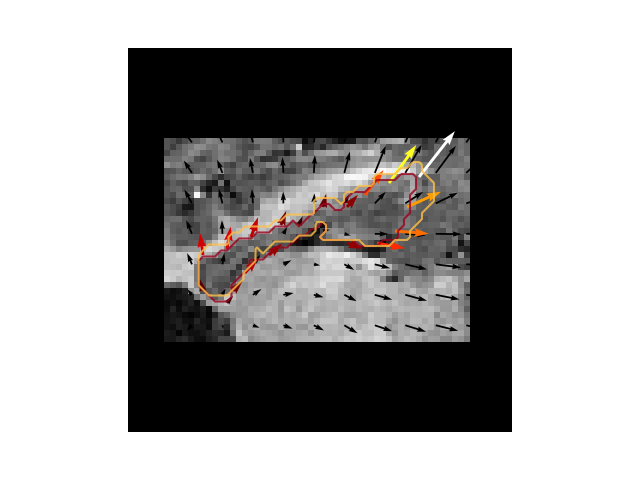}
  \end{minipage}%
  \begin{minipage}[t]{.25\textwidth}
  \centering
  \includegraphics[width=\linewidth,trim=4cm 3cm 4cm 2.8cm,clip]{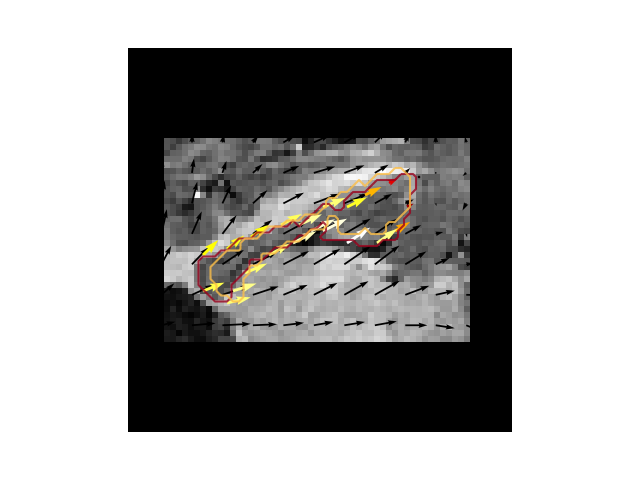}
  \end{minipage}%
  \begin{minipage}[t]{.25\textwidth}
  \centering
  \includegraphics[width=\linewidth,trim=4cm 3cm 4cm 2.8cm,clip]{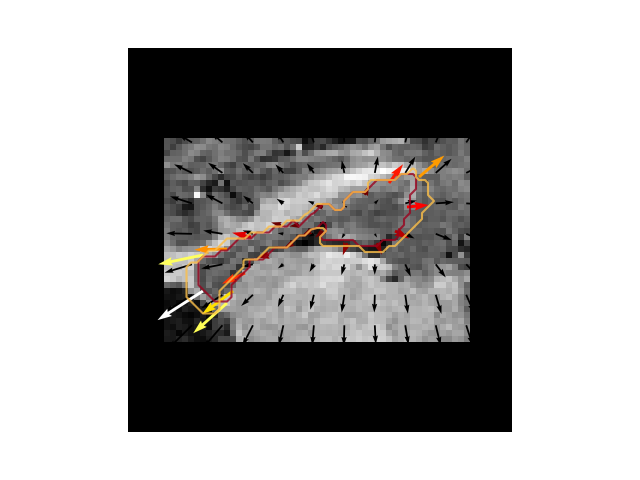}
  \end{minipage}%
  
  \caption{Extension of the Figure 4 with two other patients and the components 1 to 8 (first row 1-4, second row 5-8).}
  \label{fig:hipo_pca}
\end{figure}

\begin{figure}[h!]
    \centering
        \begin{minipage}[t]{.33\textwidth}%
        \centering
        \includegraphics[width=\linewidth,trim=0.5cm 3.4cm 2cm 2cm,clip]{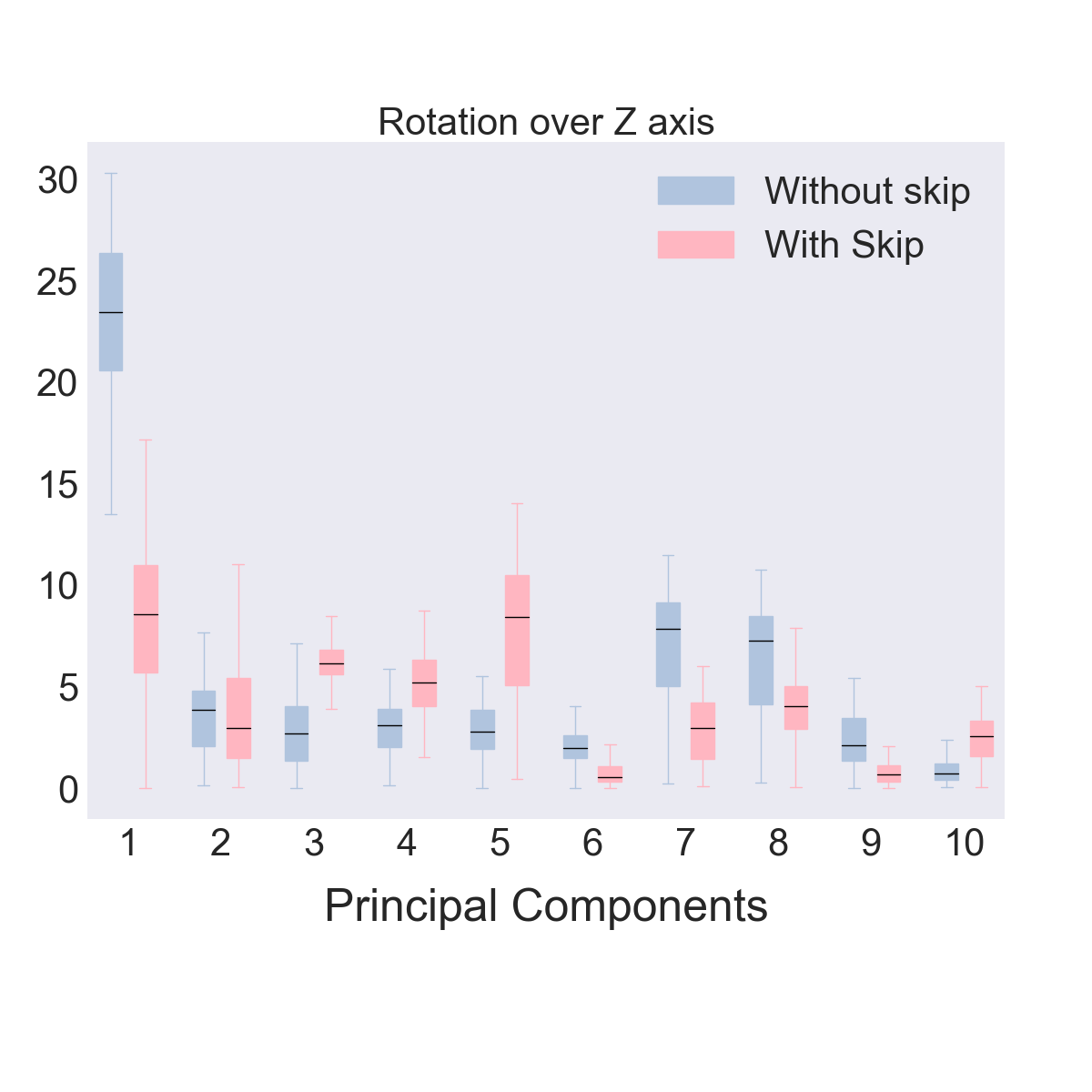}
        \end{minipage}%
        \begin{minipage}[t]{.33\textwidth}%
        \centering
        \includegraphics[width=\linewidth,trim=0.5cm 3.4cm 2cm 2cm,clip]{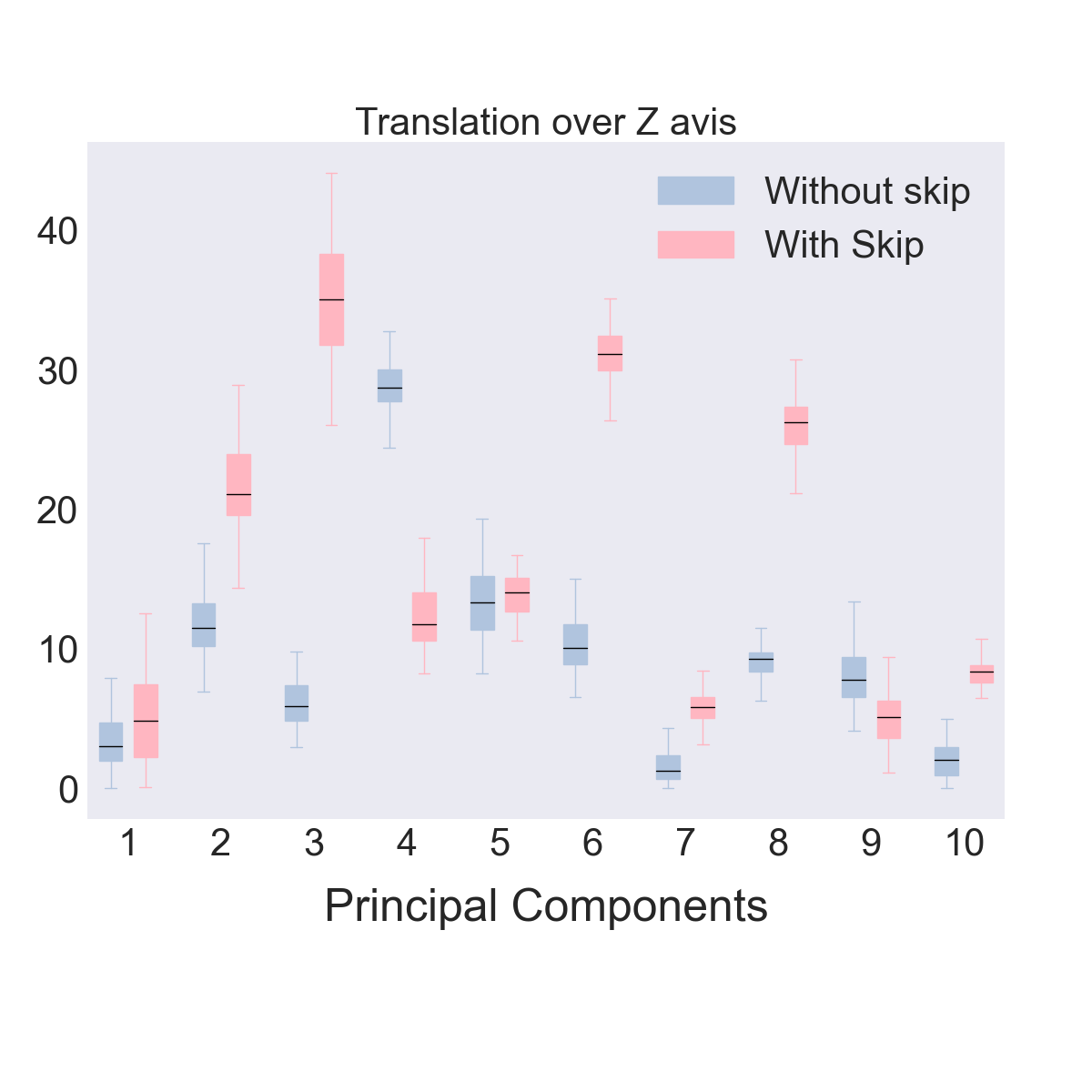}
        \end{minipage}%
        \begin{minipage}[t]{.33\textwidth}%
        \centering
        \includegraphics[width=\linewidth,trim=0.5cm 3.4cm 2cm 2cm,clip]{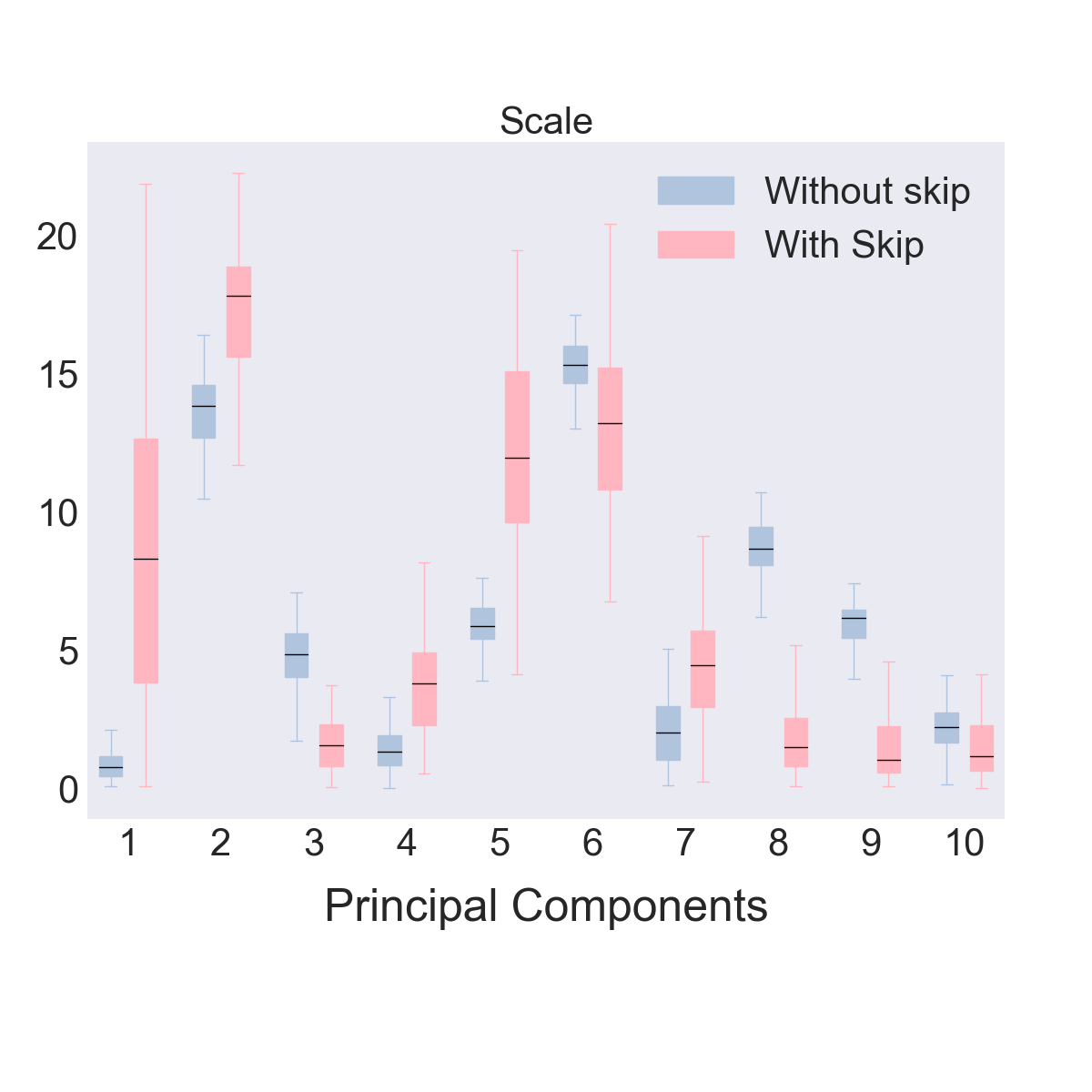}
        \end{minipage}%

    \caption{Comparison between the networks with and without skip connections. We displayed the difference between the components of an image and the same image to which we applied a predefined transformation. 
    Only one or two components are activated without the skip-connections, while many of them are with the skip.
    The first ten components have been displayed. The results are in blue and pink for respectively without and with the skip-connections. From right to left: rotation, translation along Z-axis and scaling.}
    \label{fig:comparison_skip}
\end{figure}

\end{document}